\documentclass[10pt,twocolumn,letterpaper]{article}

\usepackage{iccv}
\usepackage{times}
\usepackage{epsfig}
\usepackage{graphicx}
\usepackage{amsmath}
\usepackage{amssymb}
\usepackage{amsthm}
\usepackage{enumitem}
\usepackage{adjustbox}

\usepackage[pagebackref=true,breaklinks=true,letterpaper=true,colorlinks,bookmarks=false]{hyperref}

\usepackage{amsfonts}
\usepackage{amsmath}
\usepackage{amssymb}

\usepackage{graphicx,url}

\usepackage{xspace}
\usepackage{bm}
\usepackage[T1]{fontenc}
\usepackage{latexsym}
\usepackage{xstring}
\usepackage{relsize}

\usepackage[noend]{algorithmic}
\usepackage[vlined,ruled,linesnumbered]{algorithm2e}
\usepackage{multirow}
\usepackage{xcolor}

\usepackage{caption}
\usepackage{multicol} 

\newtheorem{theorem}{Theorem}

\newtheorem{lemma}[theorem]{Lemma}

\newtheorem{proposition}[theorem]{Proposition}
\newtheorem{remark}[theorem]{Remark}
\newtheorem{example}[theorem]{Example}

\newcommand{\bdmath}{\begin{dmath}}
\newcommand{\edmath}{\end{dmath}}
\newcommand{\beq}{\begin{equation}}
\newcommand{\eeq}{\end{equation}}
\newcommand{\bdm}{\begin{displaymath}}
\newcommand{\edm}{\end{displaymath}}
\newcommand{\bea}{\begin{eqnarray}}
\newcommand{\eea}{\end{eqnarray}}
\newcommand{\beal}{\beq \begin{array}{ll}}
\newcommand{\eeal}{\end{array} \eeq}
\newcommand{\beas}{\begin{eqnarray*}}
\newcommand{\eeas}{\end{eqnarray*}}
\newcommand{\ba}{\begin{array}}
\newcommand{\ea}{\end{array}}
\newcommand{\bit}{\begin{itemize}}
\newcommand{\eit}{\end{itemize}}
\newcommand{\ben}{\begin{enumerate}}
\newcommand{\een}{\end{enumerate}}

\newcommand{\calB}{{\cal B}}
\newcommand{\calC}{{\cal C}}

\newcommand{\calF}{{\cal F}}

\newcommand{\calL}{{\cal L}}

\newcommand{\calN}{{\cal N}}

\newcommand{\calS}{{\cal S}}

\newcommand{\calX}{{\cal X}}

\newcommand{\M}[1]{{\bm #1}} 
\renewcommand{\boldsymbol}[1]{{\bm #1}}

\newcommand{\hide}[1]{}

\newcommand{\grayout}[1]{{\color{gray} #1}}

\newcommand{\hiddenText}{{\color{gray} hidden text.}}
\newcommand{\hideWithText}[1]{\hiddenText}

\newcommand{\dist}{\mathbf{dist}}

\newcommand{\subject}{\text{ subject to }}

\DeclareMathOperator*{\argmin}{arg\,min}

\newcommand{\norm}[1]{\left\| #1 \right\|}

\newcommand{\tran}{^{\mathsf{T}}}

\newcommand{\diag}[1]{\mathrm{diag}\left(#1\right)}
\newcommand{\trace}[1]{\mathrm{tr}\left(#1\right)}

\newcommand{\inv}{^{-1}}

\newcommand{\zero}{{\mathbf 0}}
\newcommand{\eye}{{\mathbf I}}

\newcommand{\Real}[1]{ { {\mathbb R}^{#1} } }

\newcommand{\SEthree}{\ensuremath{\mathrm{SE}(3)}\xspace}

\newcommand{\SOthree}{\ensuremath{\mathrm{SO}(3)}\xspace}
\newcommand{\Othree}{\ensuremath{\mathrm{O}(3)}\xspace}

\newcommand{\MA}{\M{A}}

\newcommand{\MC}{\M{C}}

\newcommand{\MJ}{\M{J}}

\newcommand{\MM}{\M{M}}

\newcommand{\MU}{\M{U}}
\newcommand{\MR}{\M{R}}
\newcommand{\MS}{\M{S}}

\newcommand{\MV}{\M{V}}

\newcommand{\ML}{\M{L}}

\newcommand{\MT}{\M{T}}

\newcommand{\MZ}{\M{Z}}

\newcommand{\va}{\boldsymbol{a}} 
 
\newcommand{\vb}{\boldsymbol{b}}

\newcommand{\ve}{\boldsymbol{e}}
\newcommand{\vf}{\boldsymbol{f}}

\newcommand{\vn}{\boldsymbol{n}}

\newcommand{\vq}{\boldsymbol{q}}

\newcommand{\vu}{\boldsymbol{u}}
\newcommand{\vv}{\boldsymbol{v}}
\newcommand{\vt}{\boldsymbol{t}}
\newcommand{\vxx}{\boldsymbol{x}} 
\newcommand{\vy}{\boldsymbol{y}}
\newcommand{\vw}{\boldsymbol{w}}
\newcommand{\vzz}{\boldsymbol{z}}

\newcommand{\valpha}{\boldsymbol{\alpha}}

\newcommand{\vtau}{\boldsymbol{\tau}}

\newcommand{\scenario}[1]{{\smaller \sf#1}\xspace}

\newcommand{\blue}[1]{{\color{blue}#1}}

\newcommand{\linkToPdf}[1]{\href{#1}{\blue{(pdf)}}}
\newcommand{\linkToPpt}[1]{\href{#1}{\blue{(ppt)}}}
\newcommand{\linkToCode}[1]{\href{#1}{\blue{(code)}}}
\newcommand{\linkToWeb}[1]{\href{#1}{\blue{(web)}}}
\newcommand{\linkToVideo}[1]{\href{#1}{\blue{(video)}}}
\newcommand{\linkToMedia}[1]{\href{#1}{\blue{(media)}}}
\newcommand{\award}[1]{\xspace} 

\newcommand{\vz}{\boldsymbol{z}}

\newcommand{\nameshort}{\scenario{DAMP}}

\newcommand{\cbrace}[1]{\left\{#1\right\}}

\newcommand{\parentheses}[1]{\left(#1\right)}

\renewcommand{\dist}{\ensuremath{\mathrm{dist}}}
\newcommand{\pair}[1]{(#1)_{\ensuremath{\mathrm{p}}}}

\newcommand{\transform}{\otimes}

\newcommand{\calY}{\mathcal{Y}}

\renewcommand{\det}[1]{\ensuremath{\mathrm{det}}\parentheses{#1}}

\newcommand{\usphere}[1]{\mathbb{S}^{#1}}

\newcommand{\pd}[1]{\calS_{++}^{#1}}

\newcommand{\pp}{\ensuremath{\mathrm{PP}}}
\newcommand{\pl}{\ensuremath{\mathrm{PL}}}
\newcommand{\ph}{\ensuremath{\mathrm{PH}}}
\newcommand{\ps}{\ensuremath{\mathrm{PS}}}
\newcommand{\pc}{\ensuremath{\mathrm{PC}}}
\newcommand{\pk}{\ensuremath{\mathrm{PK}}}
\newcommand{\pe}{\ensuremath{\mathrm{PE}}}
\newcommand{\el}{\ensuremath{\mathrm{EL}}}

\newcommand{\hatvy}{\hat{\vy}}
\newcommand{\hatmap}[1]{[#1]_{\times}}
\newcommand{\qprod}{\odot}
\newcommand{\bmat}{\left[ \begin{array}}
\newcommand{\emat}{\end{array}\right]}

\newcommand{\ransac}{\scenario{RANSAC}}
\newcommand{\gnc}{\scenario{GNC}}
\newcommand{\bunny}{\scenario{Bunny}}
\newcommand{\speed}{\scenario{SPEED}}
\newcommand{\pascal}{\scenario{PASCAL3D+}}
\newcommand{\fgthreedcar}{\scenario{FG3DCar}}
\newcommand{\escape}{\scenario{EscapeMinimum}}
\newcommand{\true}{\scenario{True}}
\newcommand{\false}{\scenario{False}}

\newcommand{\chair}{\scenario{chair}}
\newcommand{\icp}{\scenario{ICP}}
\newcommand{\inner}[2]{\left\langle #1,#2 \right\rangle}

\newcommand{\sue}{SUE}
\newcommand{\sues}{SUEs}

\newcommand{\ux}{\underline{\vxx}}
\newcommand{\uy}{\underline{\vy}}
\newcommand{\xcm}{\vxx_{\mathrm{c}}}
\newcommand{\xref}[1]{\vxx_{\mathrm{r}_{#1}}}
\newcommand{\yref}[1]{\vy_{\mathrm{r}_{#1}}}
\newcommand{\dxcm}{\dot{\vxx}_{\mathrm{c}}}
\newcommand{\dq}{\dot{\vq}}
\newcommand{\acm}{\va_{\mathrm{c}}}
\newcommand{\accang}{\valpha}
\newcommand{\vcm}{\vv_{\mathrm{c}}}
\newcommand{\dvcm}{\dot{\vv}_{\mathrm{c}}}
\newcommand{\vomega}{\boldsymbol{\omega}}
\newcommand{\domega}{\dot{\vomega}}
\newcommand{\state}{\boldsymbol{s}}
\newcommand{\dstate}{\dot{\state}}
\newcommand{\omegahomo}{\tilde{\vomega}}
\newcommand{\barx}{\bar{\vxx}}
\newcommand{\bary}{\bar{\vy}}

\newcommand{\MUplus}{\MU_{+}}
\newcommand{\MVplus}{\MV_{+}}
\newcommand{\vtstar}{\vt^\star}
\newcommand{\MRstar}{\MR^\star}
\newcommand{\barMR}{\bar{\MR}}
\newcommand{\abs}[1]{\left|#1\right|}
\newcommand{\deltaR}{\MR_{\Delta}}
\newcommand{\deltastate}{\state_{\Delta}}

\newcommand{\supp}{Supplementary Material}

\newcommand{\errR}{e_{R}}
\newcommand{\errt}{e_{t}}
\newcommand{\diffR}{\bar{e}_{R}}
\newcommand{\difft}{\bar{e}_t}
\newcommand{\svd}{\scenario{SVD}}
\newcommand{\sdr}{\scenario{SDR}}
\newcommand{\shapestar}{\scenario{Shape$^\star$}}

\graphicspath{{./figures/}}

\iccvfinalcopy 

\def\iccvPaperID{6651} 

\ificcvfinal\fi

\begin{document}

\title{\vspace{-34mm} \small{\normalfont This paper has been accepted for publication at the International Conference on Computer Vision, 2021. Please cite the paper as: \\ H. Yang, C. Doran, and J.-J. Slotine, ``Dynamical Pose Estimation'', In \emph{International Conference on Computer Vision}, 2021.} \\ \vspace{10mm} \Large{Dynamical Pose Estimation} \vspace{-4mm}}

\author{Heng Yang\\
MIT
\and
Chris Doran\\
University of Cambridge
\and
Jean-Jacques Slotine \\
MIT
}

\twocolumn[{%
\renewcommand\twocolumn[1][]{#1}%
\maketitle
\vspace{-8mm}
\newcommand{\mpwfive}{3.4cm}
\begin{minipage}{\textwidth}
	\begin{center}
	\begin{minipage}{\textwidth}
	\begin{tabular}{ccccc}%
		\hspace{-6mm}	\begin{minipage}{\mpwfive}%
			\centering%
			\includegraphics[width=\columnwidth]{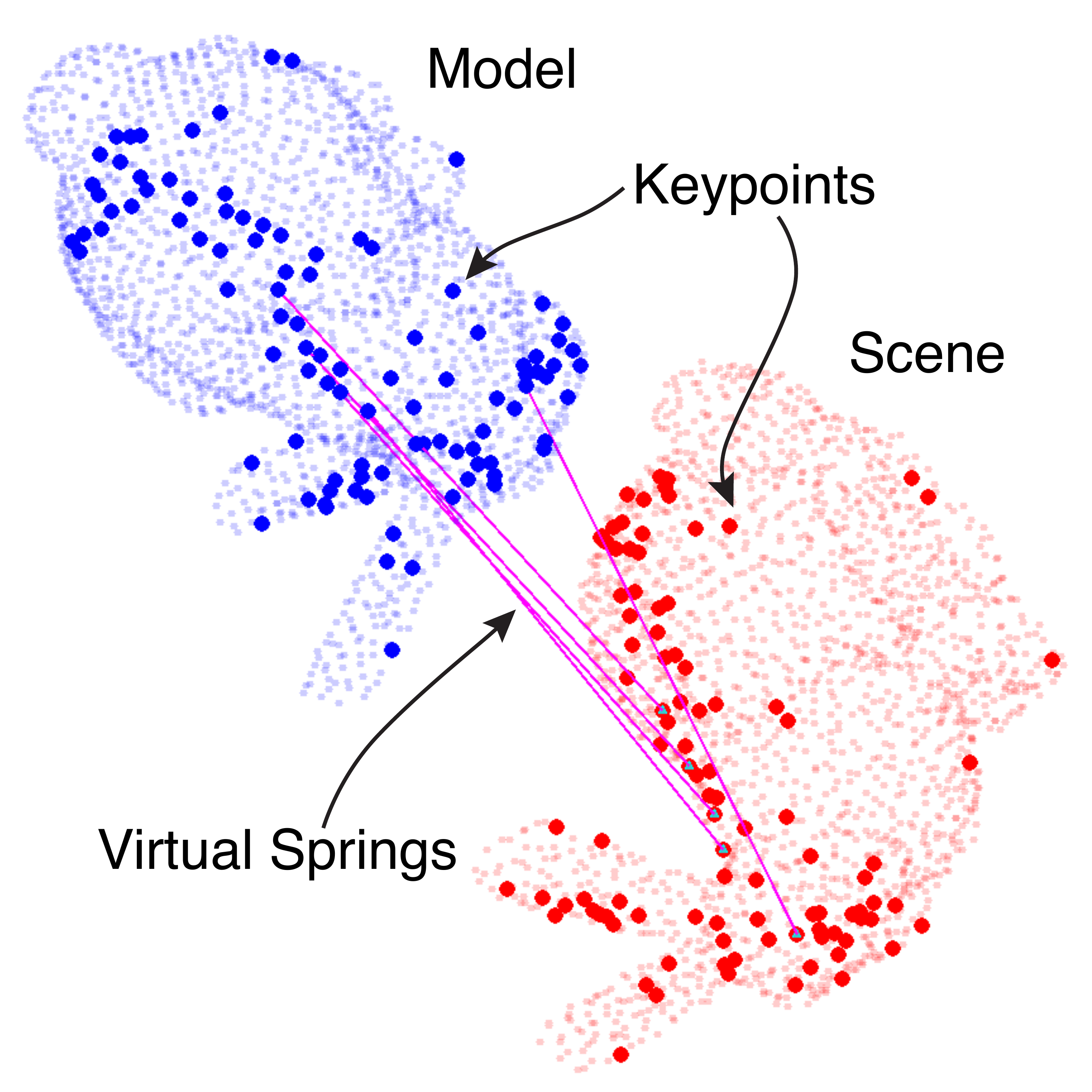}
			\end{minipage}
		&
			\begin{minipage}{\mpwfive}%
			\centering%
			\includegraphics[width=\columnwidth]{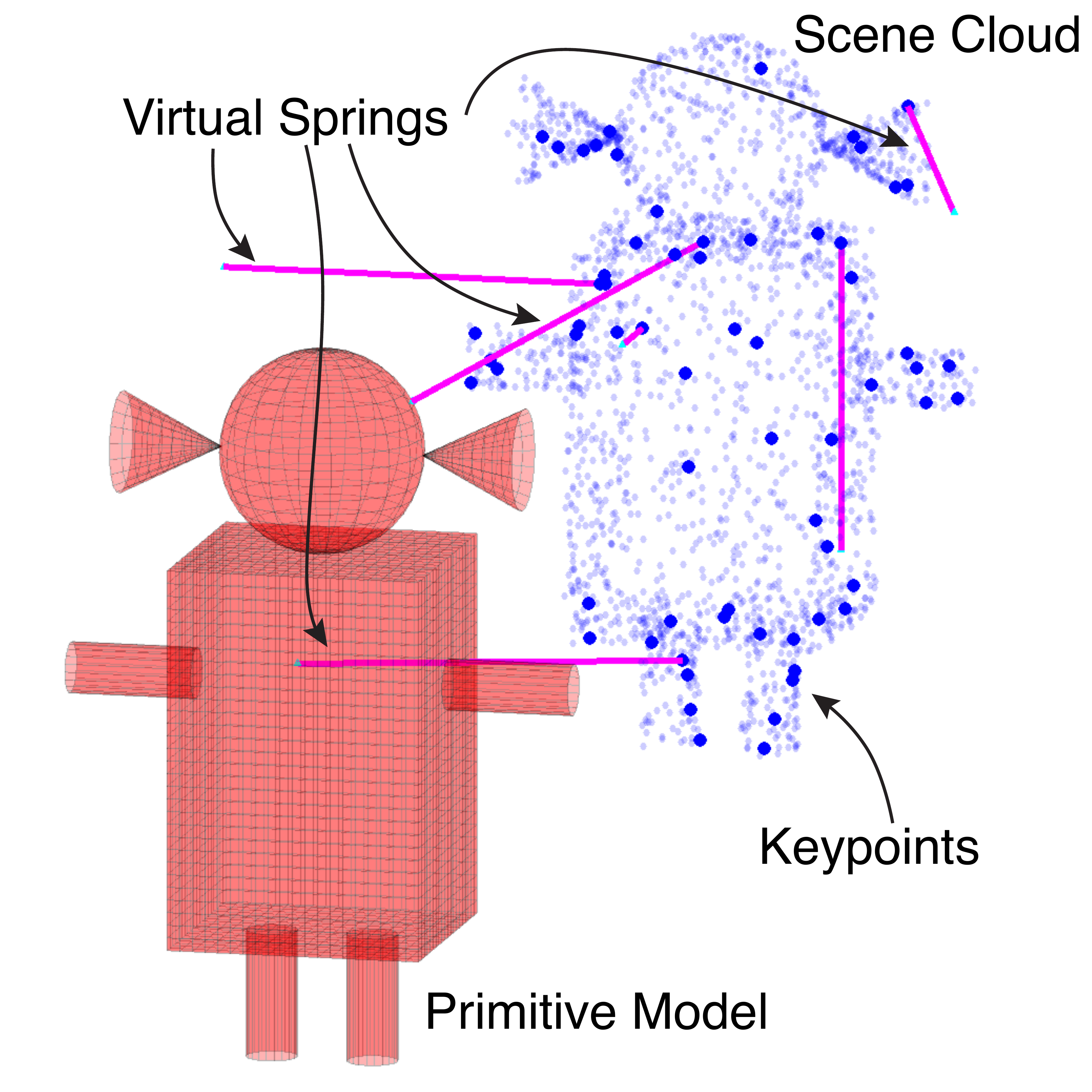}
			\end{minipage}
		& 
			\begin{minipage}{\mpwfive}%
			\centering%
			\includegraphics[width=\columnwidth]{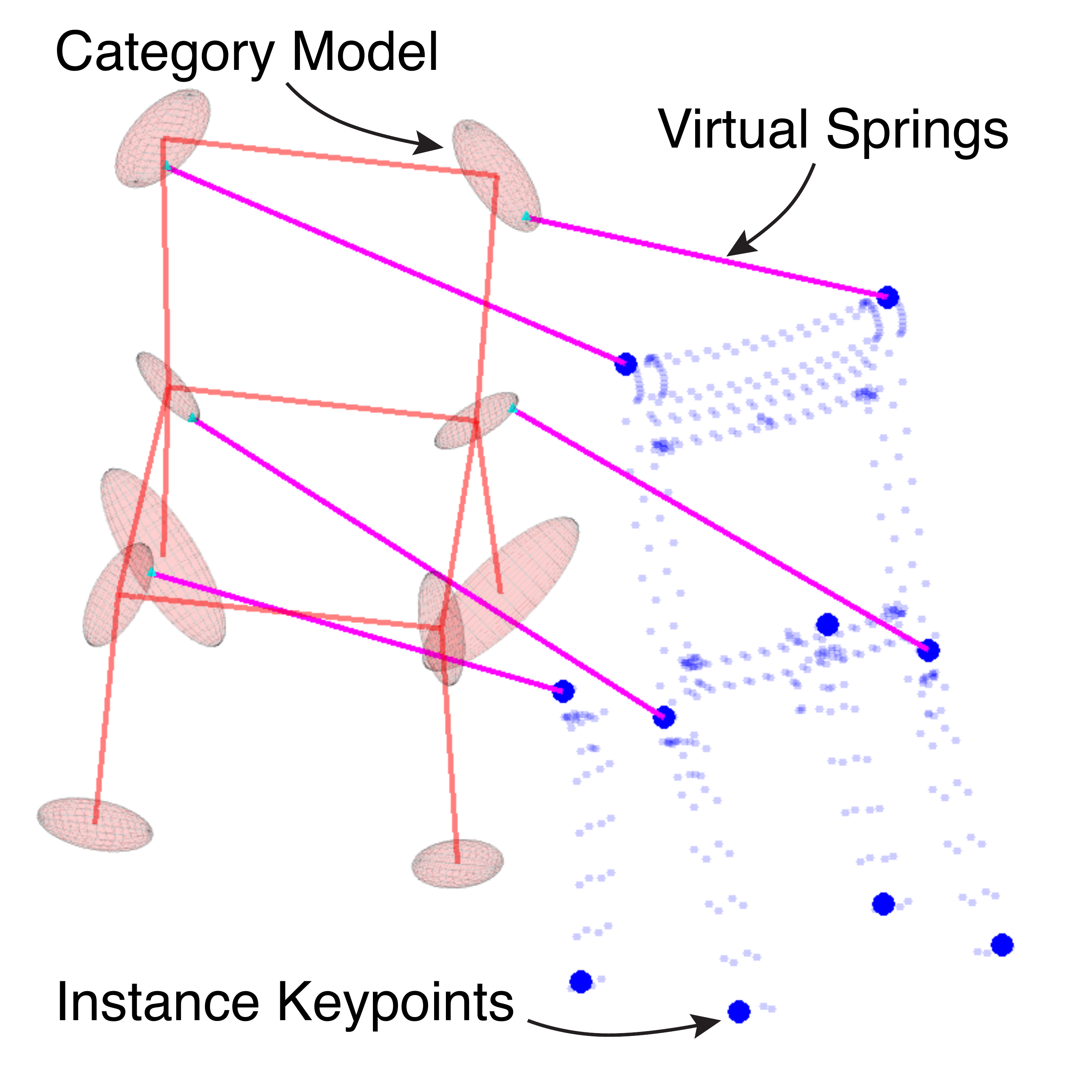}
			\end{minipage}
		& 
			\begin{minipage}{\mpwfive}%
			\centering%
			\includegraphics[width=\columnwidth]{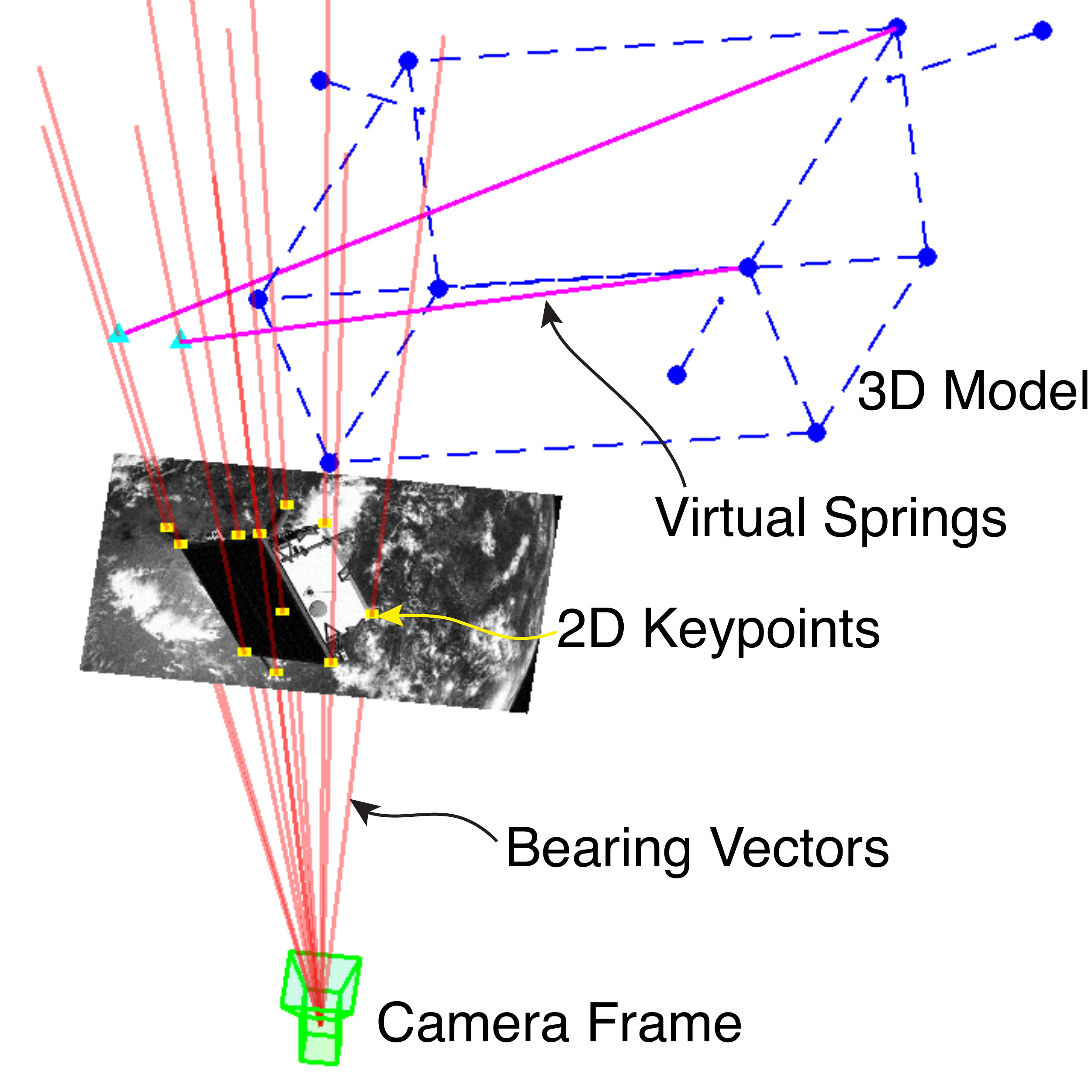}
			\end{minipage}
		& 
			\begin{minipage}{\mpwfive}%
			\centering%
			\includegraphics[width=\columnwidth]{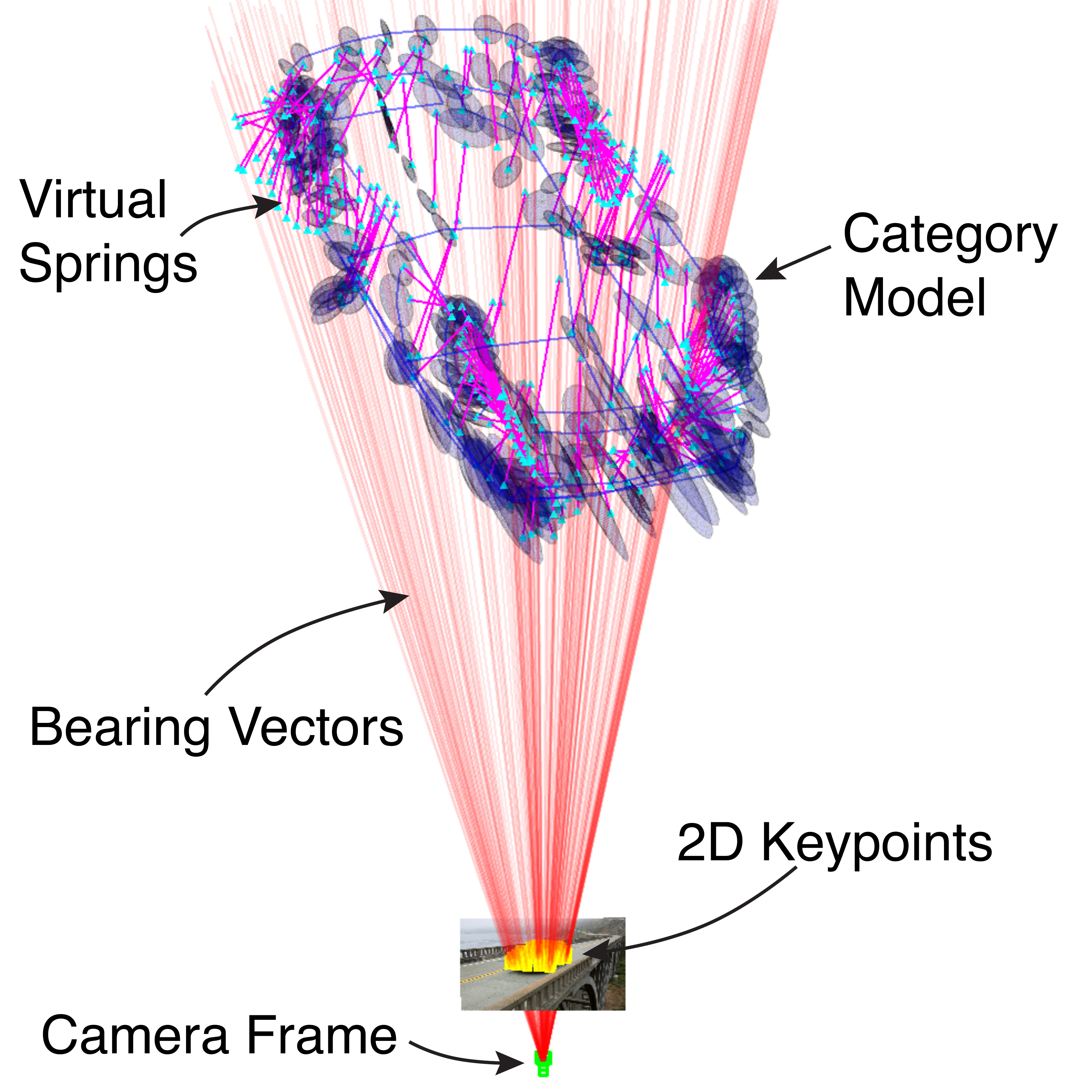}
			\end{minipage}
		\\

		\hspace{-8mm}	\begin{minipage}{\mpwfive}%
			\centering%
			\includegraphics[width=\columnwidth]{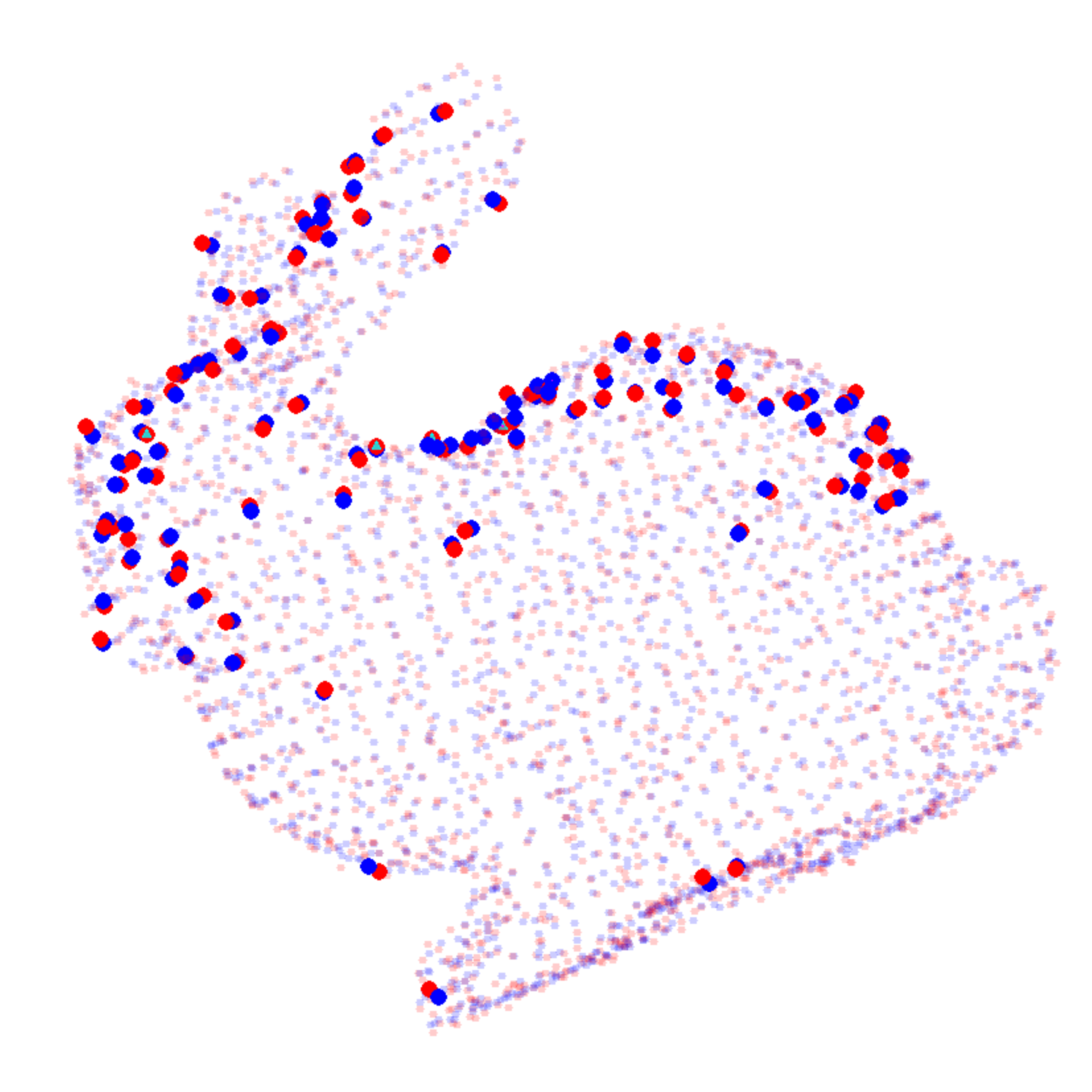} \\
			{\smaller (a) Point Cloud Registration}
			\end{minipage}
		&
			\begin{minipage}{\mpwfive}%
			\centering%
			\includegraphics[width=\columnwidth]{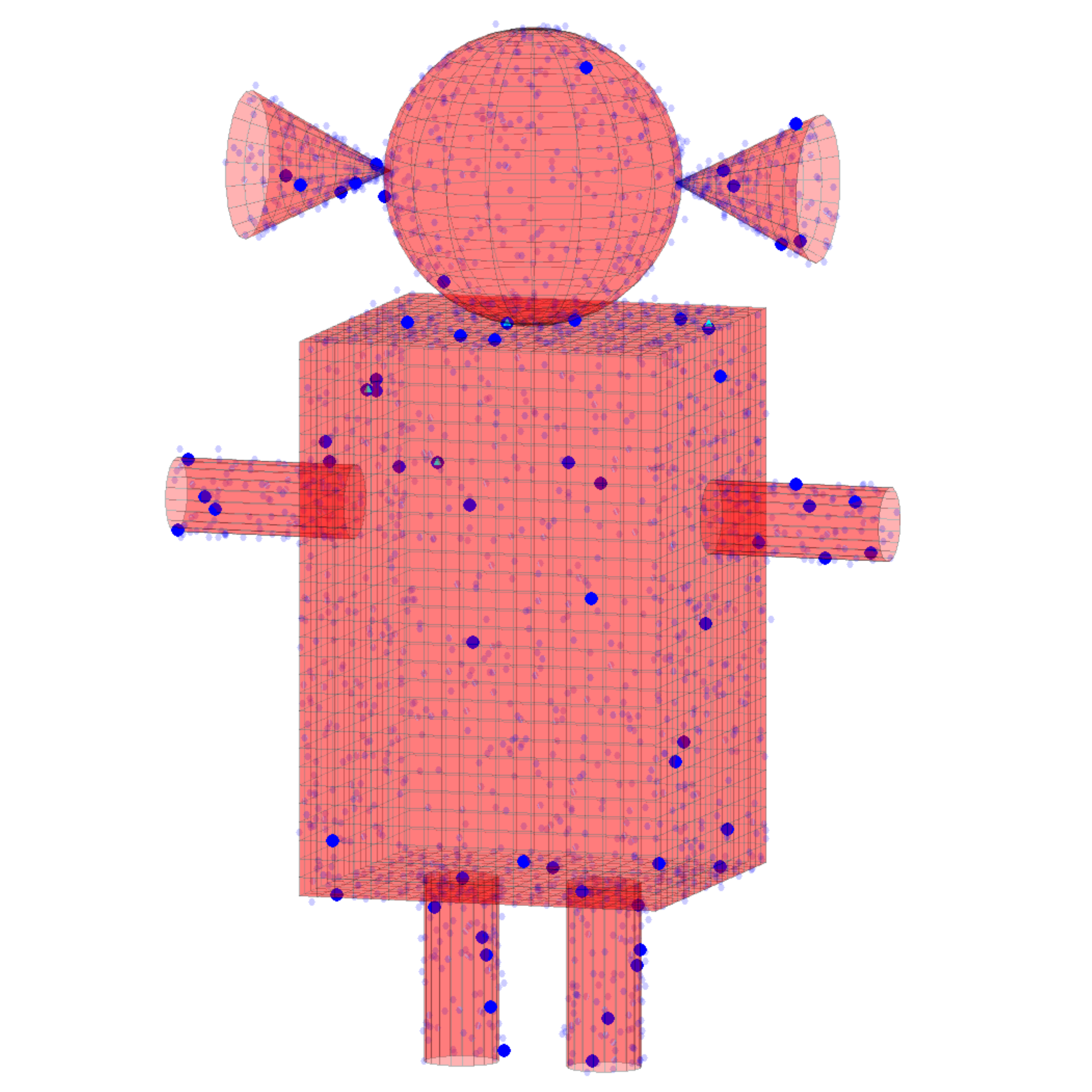}\\
			{\smaller (b) Primitive Registration}
			\end{minipage}
		& 
			\begin{minipage}{\mpwfive}%
			\centering%
			\includegraphics[width=\columnwidth]{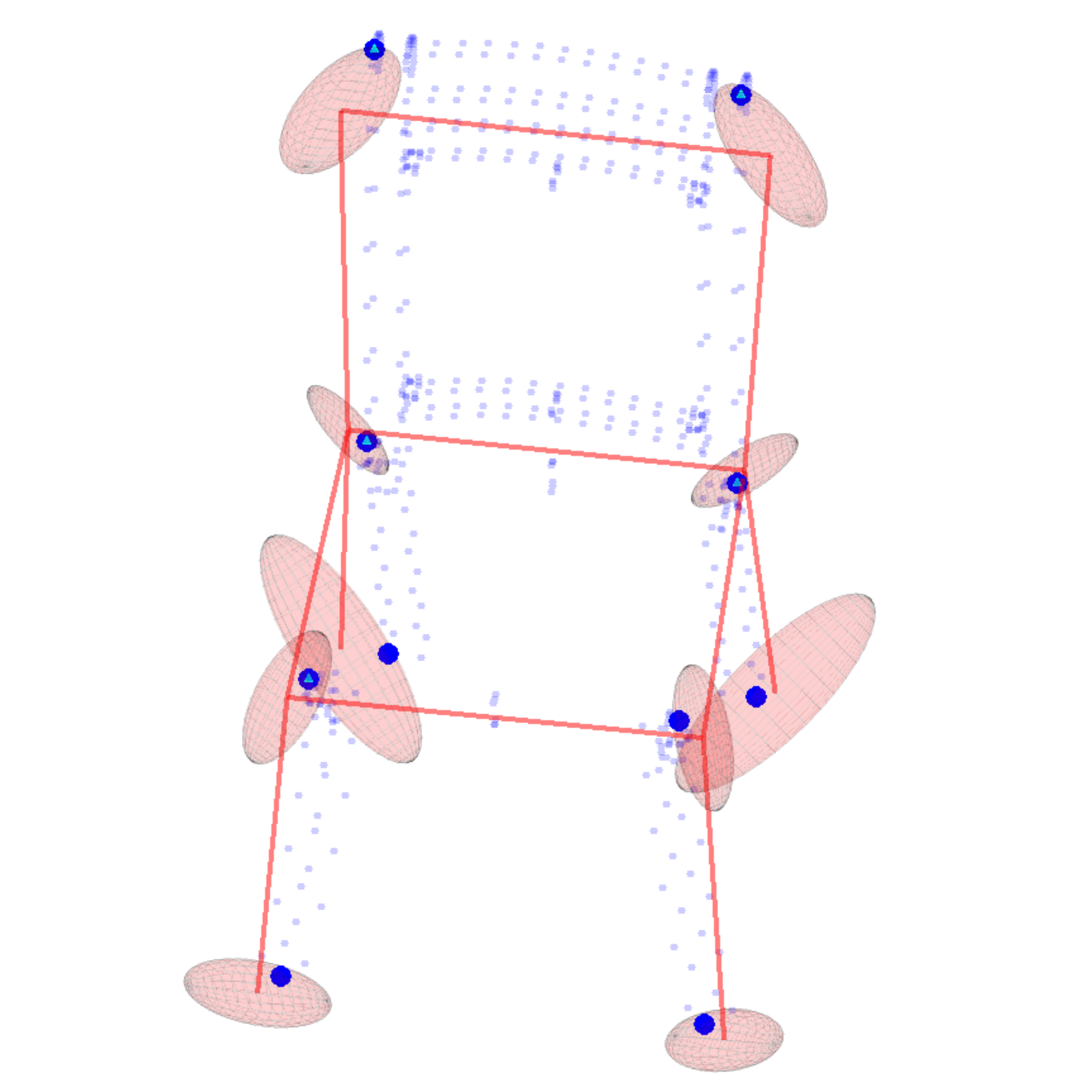}\\
			{\smaller (c) Category Registration}
			\end{minipage}
		& 
			\begin{minipage}{\mpwfive}%
			\centering%
			\includegraphics[width=\columnwidth]{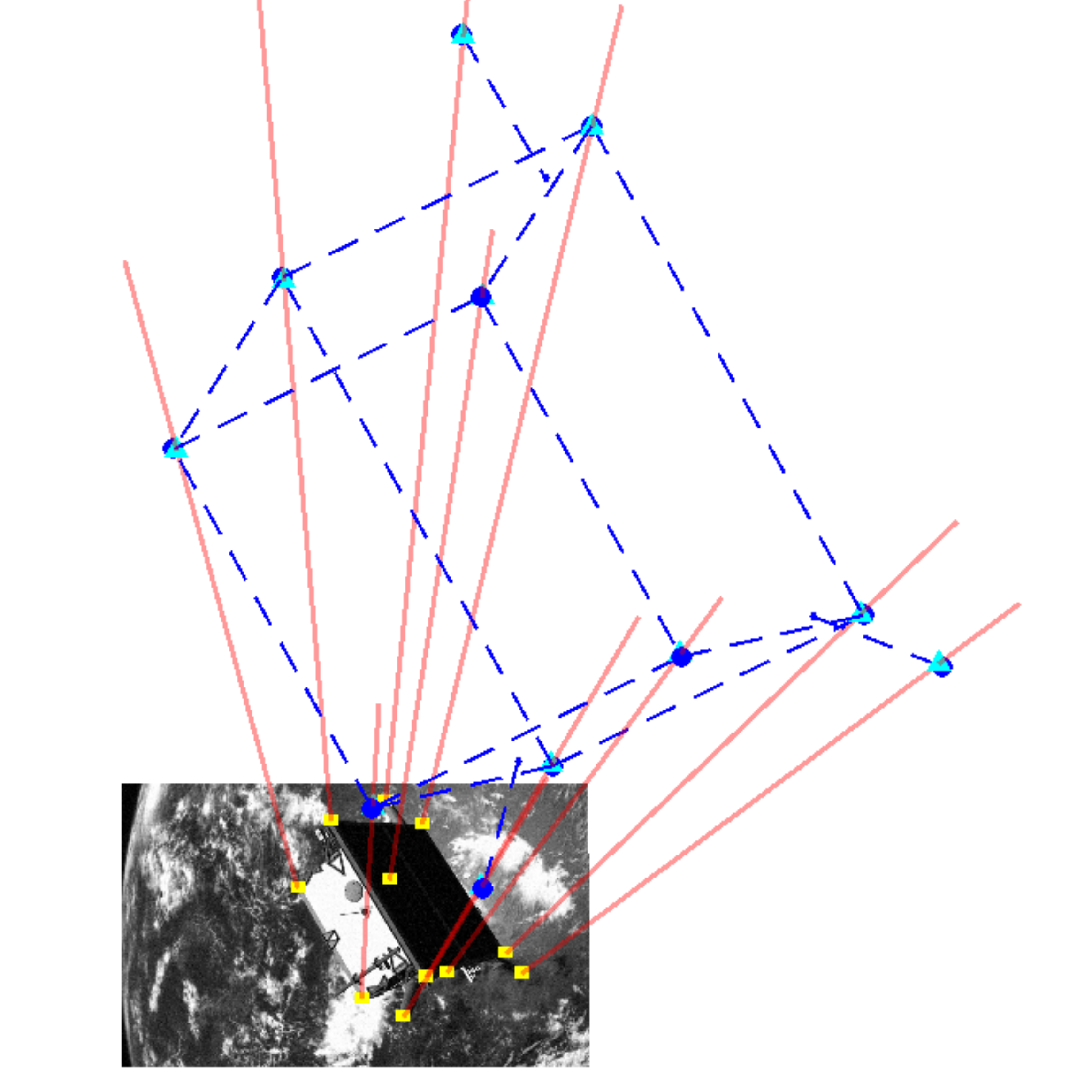}\\
			{\smaller (d) Absolute Pose Estimation}
			\end{minipage}
		& 
			\begin{minipage}{\mpwfive}%
			\centering%
			\includegraphics[width=\columnwidth]{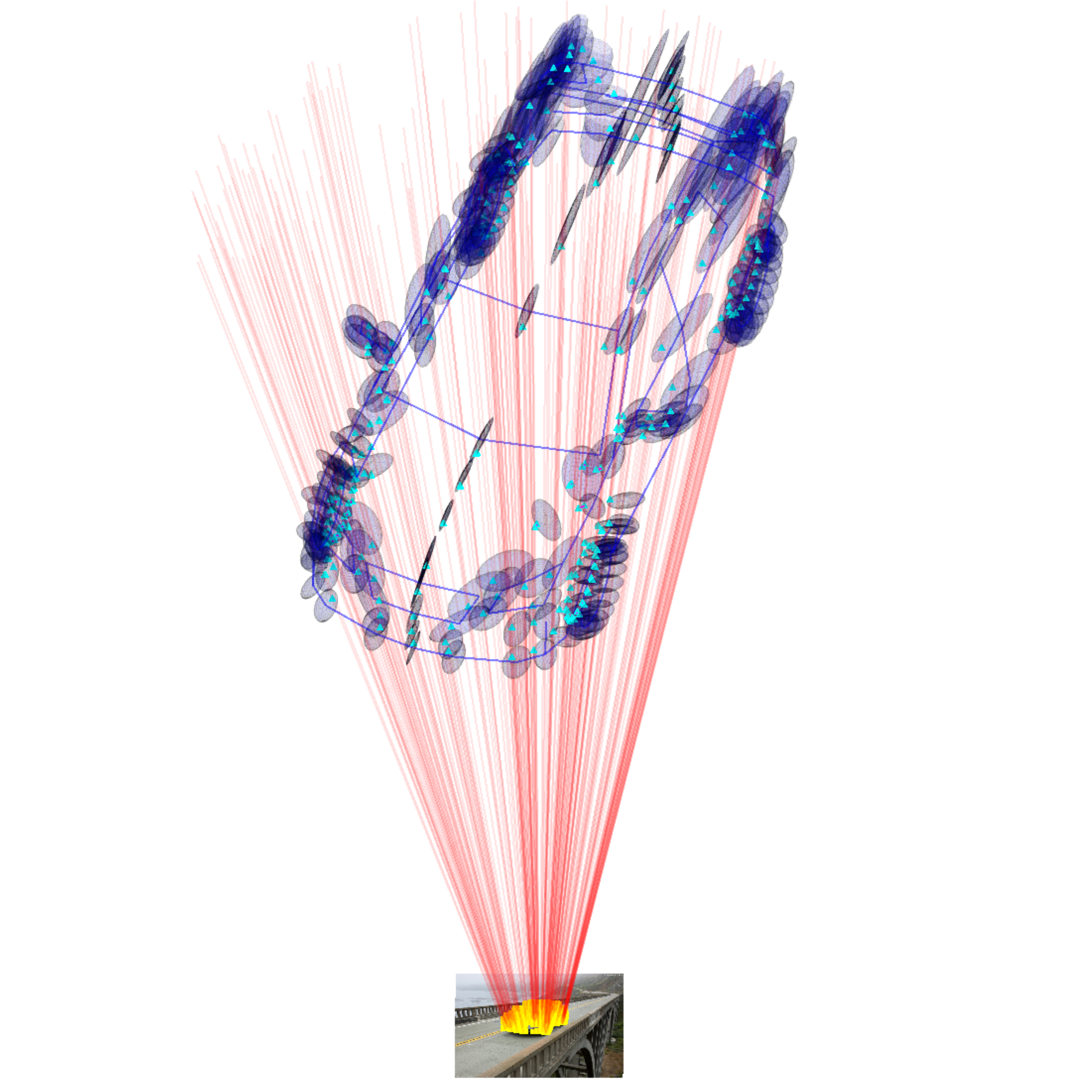}\\
			{\smaller (e) Category APE}
			\end{minipage}
	\end{tabular}
	\vspace{-3mm}
	\end{minipage} 
	\captionof{figure}{We propose \emph{DynAMical Pose estimation} (\nameshort), the first general and practical framework to perform pose estimation from 2D and 3D visual correspondences by simulating \emph{rigid body dynamics} arising from \emph{virtual springs and damping} (top row, magenta lines). \nameshort~almost always returns the \emph{globally} optimal rigid transformation across five pose estimation problems (bottom row). (a) Point cloud registration using the \bunny~dataset~\cite{Curless96siggraph}; (b) Primitive registration using a robot model of spheres, planes, cylinders and cones; (c) Category registration using the~\chair~category from the \pascal~dataset~\cite{Xiang2014WACV-PASCAL+}; (d) Absolute pose estimation (APE) using the \speed~satellite dataset~\cite{Sharma19arXiv-SPEED};  (e) Category APE using the \fgthreedcar~dataset~\cite{Lin14eccv-modelFitting}. \vspace{-2mm}
	\label{fig:all-apps}} 
	\end{center}
\end{minipage}

}]


\begin{abstract}
\vspace{-4mm}
We study the problem of aligning two sets of 3D geometric primitives given known correspondences. Our first contribution is to show that this \emph{primitive alignment} framework unifies five perception problems including point cloud registration, primitive (mesh) registration, category-level 3D registration, absolution pose estimation (APE), and category-level APE. Our second contribution is to propose DynAMical Pose estimation (\nameshort), the first general and practical algorithm to solve primitive alignment problem by simulating \emph{rigid body dynamics} arising from virtual springs and damping, where the springs span the \emph{shortest distances} between corresponding primitives. We evaluate \nameshort in simulated and real datasets across all five problems, and demonstrate (i) \nameshort always converges to the \emph{globally optimal} solution in the first three problems with 3D-3D correspondences; (ii) although \nameshort sometimes converges to suboptimal solutions in the last two problems with 2D-3D correspondences, using a scheme for escaping local minima, \nameshort always succeeds. Our third contribution is to demystify the surprising empirical performance of \nameshort and formally prove a global convergence result in the case of point cloud registration by charactering local stability of the equilibrium points of the underlying dynamical system.\footnote{Code: \url{https://github.com/hankyang94/DAMP}}
\end{abstract}

\ificcvfinal\thispagestyle{empty}\fi

\vspace{-10mm}
\section{Introduction}
Consider the problem of finding the best \emph{rigid} transformation (pose) to align two sets of \emph{corresponding} 3D geometric \emph{primitives} $\calX = \cbrace{X_i}_{i=1}^N$ and $\calY = \cbrace{Y_i}_{i=1}^N$:
\bea
\min_{\MT \in \SEthree} \sum_{i=1}^N  \dist\parentheses{\MT \transform X_i, Y_i}^2, \label{eq:primitivealignment}
\eea
where $\SEthree \triangleq \cbrace{(\MR,\vt): \MR \in \SOthree, \vt \in \Real{3}}$\footnote{$\SOthree \triangleq \cbrace{\MR \in \Real{3 \times 3}: \MR\MR\tran = \MR\tran\MR = \eye_3, \det{\MR} = +1}$ is the set of proper 3D rotations.} is the set of 3D rigid transformations (rotations and translations), $\MT \transform X$ denotes the action of a rigid transformation $\MT$ on the primitive $X$, and $\dist\parentheses{X,Y}$ is the \emph{shortest} distance between two primitives $X$ and $Y$. In particular, we focus on the following primitives:
\begin{enumerate}
	\itemsep-0.2em
	\item \label{item:point} \emph{Point}: $P(\vxx) \triangleq \cbrace{\vxx}$, where $\vxx \in \Real{3}$ is a 3D point;

	\item \label{item:line} \emph{Line}: $L(\vxx,\vv) \triangleq \cbrace{\vxx + \alpha \vv: \alpha \in \Real{}}$, where $\vxx \in \Real{3}$ is a point on the line, and $\vv \in \usphere{2}$ is the unit direction;\footnote{$\usphere{n-1} \triangleq \cbrace{\vv \in \Real{n}: \norm{\vv} = 1}$ is the set of $n$-D unit vectors.}

	\item \label{item:plane} \emph{Plane}: $H(\vxx,\vn)\triangleq \cbrace{\vy \in \Real{3}: \vn\tran (\vy - \vxx) = 0}$, where $\vxx \in \Real{3}$ is a point on the plane, and $\vn \in \usphere{2}$ is the unit normal that is perpendicular to the plane;

	\item \label{item:sphere} \emph{Sphere}: $S(\vxx,r) \triangleq \{\vy \in \Real{3}: \norm{\vy - \vxx}^2 = r^2 \}$, where $\vxx \in \Real{3}$ is the center, and $r>0$ is the radius; 

	\item \label{item:cylinder} \emph{Cylinder}: $C(\vxx,\vv,r) \triangleq \{\vy \in \Real{3}:\dist(\vy,L(\vxx,\vv)) = r\}$, where $L(\vxx,\vv)$ (defined in~\ref{item:line}) is the central axis of the cylinder, $r>0$ is the radius, and $\dist(\vy,L)$ is the orthogonal distance from point $\vy$ to line $L$;

	\item \label{item:cone} \emph{Cone}: $K(\vxx,\vv,\theta)\!\triangleq\!\{\vy\!\in\!\Real{3}:\!\vv\tran (\vy - \vxx)\!=\!\cos\theta\!\norm{\vy\!-\!\vxx}\}$, where $\vxx \in \Real{3}$ is the apex, $\vv \in \usphere{2}$ is the unit direction of the central axis pointing inside the cone, and $\theta \in (0,\frac{\pi}{2})$ is the half angle; 

	\item \label{item:ellipsoid} \emph{Ellipsoid}: $E(\vxx,\!\MA)\!\!\triangleq\!\!\cbrace{\vy\!\in\!\Real{3}\!:\!\!(\vy\!-\!\vxx)\tran \MA (\vy\!-\!\vxx)\!\leq \!1}$, where $\vxx \in \Real{3}$ is the center, and $\MA \in \pd{3}$ is a positive definite matrix defining the principal axes.\footnote{$\calS^n, \calS_{+}^n, \calS_{++}^n$ denote the set of real $n \times n$ symmetric, positive semidefinite, and positive definite matrices, respectively.}
\end{enumerate}

Problem~\eqref{eq:primitivealignment}, when specialized to the primitives~\ref{item:point}-\ref{item:ellipsoid}, includes a broad class of fundamental perception problems concerning \emph{pose estimation} from visual measurements, and finds extensive applications to object detection and localization~\cite{Kneip2014ECCV-UPnP,Peng19CVPR-PVNet}, motion estimation and 3D reconstruction~\cite{Yang20tro-teaser,Yang20neurips-certifiablePerception}, and simultaneous localization and mapping~\cite{Cadena16tro-SLAMsurvey,Yang20ral-GNC,Rosinol20icra-Kimera}.

In this paper, we consider five examples of problem~\eqref{eq:primitivealignment}, with graphical illustrations given in Fig.~\ref{fig:all-apps}. Note that we restrict ourselves to the case when all correspondences $X_i \leftrightarrow Y_i,i=1,\dots,N$, are \emph{known and correct}, for two reasons: (i) there are general-purpose algorithmic frameworks, such as \ransac~\cite{Fischler81} and \gnc~\cite{Yang20ral-GNC,Antonante20arxiv-outlierRobustEstimation} that re-gain robustness to incorrect correspondences (\ie~\emph{outliers}) once we have efficient solvers for the outlier-free problem~\eqref{eq:primitivealignment}; (ii) even when all correspondences are correct, problem~\eqref{eq:primitivealignment} can be difficult to solve due to the non-convexity of the feasible set $\SEthree$.

\begin{example}[Point Cloud Registration~\cite{Horn87josa,Yang19rss-teaser}]
\label{ex:pointcloudregistration}
Let $X_i = P(\vxx_i)$ and $Y_i = P(\vy_i)$ in problem~\eqref{eq:primitivealignment}, with $\vxx_i,\vy_i \in \Real{3}$, point cloud registration seeks the best rigid transformation to align two sets of 3D points.
\end{example}
Fig.~\ref{fig:all-apps}(a) shows an instance of point cloud registration using the \bunny dataset~\cite{Curless96siggraph}, with bold blue and red dots being the \emph{keypoints} $P(\vxx_i)$ and $P(\vy_i)$, respectively. Point cloud registration commonly appears when one needs to align two or more Lidar or RGB-D scans acquired at different space and time~\cite{Yang20tro-teaser}, and in practice either hand-crafted~\cite{Rusu09icra-fast3Dkeypoints} or deep-learned~\cite{Choy19iccv-FCGF,Gojcic19cvpr-3Dsmoothnet,Yang21cvpr-SGP} feature descriptors are adopted to generate point-to-point correspondences. 

However, in many cases it is challenging to obtain (in run time), or annotate (in training time), point-to-point correspondences (\eg,~it is much easier to tell a point lies on a plane than to precisely localize where it lies on the plane as in Fig.~\ref{fig:all-apps}(b)). Moreover, it is well known that correspondences such as point-to-line and point-to-plane ones can lead to better convergence in algorithms such as \icp~\cite{Besl92pami}. Recently, a growing body of research seeks to represent and approximate complicated 3D shapes using simple primitives such as cubes, cones, cylinders etc.~to gain efficiency in storage and capability in assigning semantic meanings to different parts of a 3D shape~\cite{Tulsiani17cvpr-shapeabstract,Genova19iccv-shapetemplate,li19cvpr-primitiveFitting}. These factors motivate the following primitive registration problem.

\begin{example}[Primitive Registration~\cite{Briales17cvpr-registration,li19cvpr-primitiveFitting}]
\label{ex:primitiveRegistration}
Let $X_i=P(\vxx_i), \vxx \in \Real{3}$, be a 3D point, and let $Y_i$ be any type of primitives among~\ref{item:point}-\ref{item:ellipsoid} in problem~\eqref{eq:primitivealignment}, primitive registration seeks the best rigid transformation to align a set of 3D points to a set of 3D primitives. 
\end{example}

Fig.~\ref{fig:all-apps}(b) shows an example where a semantically meaningful robot model is compactly represented as a collection of planes, cylinders, spheres and cones, while a noisy point cloud observation is aligned to it by solving problem~\eqref{eq:primitivealignment}.

Both Examples~\ref{ex:pointcloudregistration} and~\ref{ex:primitiveRegistration} require a known 3D model, either in the form of a clean point cloud or a collection of fixed primitives, which can be quite restricted. For example, in Fig.~\ref{fig:all-apps}(c), imagine a robot has seen multiple \emph{instances} of a chair and only stored a \emph{deformable} model (shown in red) of the category ``\emph{chair}'' in the form of a collection of \emph{semantic uncertainty ellipsoids} (\sue), where the center of each ellipsoid keeps the \emph{average} location of a semantic keypoint (\eg,~legs of a chair) while the orientation and size of the ellipsoid represent \emph{intra-class variations} of that keypoint within the category (see \supp~for details about how \sues~are computed from data). Now the robot sees an instance of a chair (shown in blue) that either it has never seen before, or it has seen but does not have access to a precise 3D model, and has to estimate the pose of the instance \wrt~itself. In this situation, we formulate a \emph{category-level 3D registration} using \sues.

\begin{example}[Category Registration~\cite{Manuelli19isrr-kpam,Chabot17cvpr-deepManta,Shi21rss-pace}]
\label{ex:categoryregistration}
Let $X_i = P(\vxx_i), \vxx_i \in \Real{3}$, be a 3D point, and $Y_i = E(\vy_i,\MA_i), \vy_i \in \Real{3},\MA_i \in \pd{3}$, be a \sue~of a semantic keypoint, category registration seeks the best rigid transformation to align a point cloud to a set of category-level semantic keypoints.
\end{example}

The above three Examples~\ref{ex:pointcloudregistration}-\ref{ex:categoryregistration} demonstrate the flexibility of problem~\eqref{eq:primitivealignment} in modeling pose estimation problems given 3D-3D correspondences. The next two examples show that pose estimation given 2D-3D correspondences (\ie,~\emph{absolute pose estimation} (APE) or \emph{perspective-$n$-points} (PnP)) can also be formulated in the form of problem~\eqref{eq:primitivealignment}. The crux is the insight that a 2D image keypoint is uniquely determined (assume camera intrinsics are known) by a so-called \emph{bearing vector} that originates from the camera center and goes through the 2D keypoint on the imaging plane (\cf~Fig.~\ref{fig:all-apps}(d))~\cite{Hartley04book-multiviewgeometry}.\footnote{Similarly, a 2D line on the imaging plane can be uniquely determined by a 3D plane containing two bearing vectors that intersects two 2D points on the imaging plane. Therefore, problem~\eqref{eq:primitivealignment} can also accommodate point-to-line correspondences commonly seen in the literature of perspective-$n$-points-and-lines (PnPL)~\cite{Agostinho2019arXiv-cvxpnpl,Liu20ral-BnBPnL}.} Consequently, APE can be formulated as aligning the 3D model to a set of 3D bearing vectors.

\begin{example}[Absolute Pose Estimation~\cite{Kneip2014ECCV-UPnP,Agostinho2019arXiv-cvxpnpl}]
\label{ex:absolutepose}
Let $X_i = P(\vxx_i), \vxx_i \in \Real{3}$, be a 3D point, and $Y_i = L(\zero,\vv_i)$, $\vv_i \in \usphere{2}$, be the bearing vector of a 2D keypoint (the camera center is $\zero \in \Real{3}$), APE seeks to find the best rigid transformation to align a 3D point cloud to a set of bearing vectors.
\end{example}

Fig.~\ref{fig:all-apps}(d) shows an example of aligning a satellite wireframe model to a set of 2D keypoint detections. Similarly, by allowing the 3D model to be a collection of \sues, we can generalize Example~\ref{ex:absolutepose} to category-level APE.

\begin{example}[Category Absolute Pose Estimation~\cite{Yang20cvpr-shapeStar,Lin14eccv-modelFitting}]
\label{ex:categoryabsolutepose}
Let $X_i = E(\vxx_i,\MA_i)$, $\vxx_i \in \Real{3}, \MA_i \in \pd{3}$, be a \sue~of a category-level semantic keypoint, and $Y_i = L(\zero,\vv_i)$, $\vv_i \in \usphere{2}$, be the bearing vector of a 2D keypoint, category APE seeks to find the best rigid transformation to align a 3D category to the 2D keypoints of an instance.
\end{example}

Fig.~\ref{fig:all-apps}(e) shows an example of estimating the pose of a car using a category-level collection of \sues. Strictly speaking, Example~\ref{ex:primitiveRegistration} contains Examples~\ref{ex:pointcloudregistration}, \ref{ex:categoryregistration} and~\ref{ex:absolutepose}, but we separate them because they have different applications. 

{\bf Related Work}. To the best of our knowledge, this is the first time that the five seemingly different examples are formulated under the same framework. We shall briefly discuss existing methods for solving them. Point cloud registration (Example~\ref{ex:pointcloudregistration}) can be solved in closed form using singular value decomposition~\cite{Horn87josa,Arun87pami}. A comprehensive review of recent advances in point cloud registration, especially on dealing with outliers, can be found in~\cite{Yang20tro-teaser}. The other four examples, however, do not admit closed-form solutions. Primitive registration (Example~\ref{ex:primitiveRegistration}) in the case of point-to-point, point-to-line and point-to-plane correspondences (referred to as \emph{mesh registration}~\cite{Yang20neurips-certifiablePerception}) can be solved globally using branch-and-bound~\cite{Olsson09pami-bnbRegistration} and semidefinite relaxations~\cite{Briales17cvpr-registration}, hence, is relatively slow. Further, there are no solvers that can solve primitive registration including point-to-sphere, point-to-cylinder and point-to-cone correspondences with global optimality guarantees. The absolute pose estimation problem (Example~\ref{ex:absolutepose}) has been a major line of research in computer vision, and there are several global solvers based on Grobner bases~\cite{Kneip2014ECCV-UPnP} and convex relaxations~\cite{Agostinho2019arXiv-cvxpnpl,Schweighofer2008bmvc-SOSforPnP}. For category-level registration and APE (Example~\ref{ex:categoryregistration} and~\ref{ex:categoryabsolutepose}), most existing methods formulate them as simultaneously estimating the \emph{shape coefficients} and the camera pose,~\ie,~they treat the unknown instance model as a \emph{linear combination} of category templates (known as the \emph{active shape model}~\cite{Cootes95cviu}) and seek to estimate the linear coefficients as well as the camera pose. Works in~\cite{Gu06cvpr-faceAlignment,Ramakrishna12eccv-humanPose,Lin14eccv-modelFitting} solve the joint optimization by alternating the estimation of the shape coefficients and the estimation of the camera pose, thus requiring a good initial guess for convergence. Zhou~\etal~\cite{Zhou15cvpr,Zhou17pami-shapeEstimationConvex} developed a convex relaxation technique to solve category APE with a \emph{weak perspective} camera model and showed efficient and accurate results. Yang and Carlone~\cite{Yang20cvpr-shapeStar} later showed that the convex relaxation in~\cite{Zhou15cvpr,Zhou17pami-shapeEstimationConvex} is less tight than the one they developed based on sums-of-squares (SOS) relaxations. However, the SOS relaxation in~\cite{Yang20cvpr-shapeStar} leads to large semidefinite programs (SDP) that cannot be solved efficiently at present time. Very recently, with the advent of machine learning, many researchers resort to deep networks that regress the 3D shape and the camera pose directly from 2D images~\cite{Chabot17cvpr-deepManta,Kolotouros19iccv-humanlearnplusmodel,Tatarchenko19CVPR-singleViewReconLimitation}. We refer the interested reader to~\cite{Tatarchenko19CVPR-singleViewReconLimitation,Kolotouros19iccv-humanlearnplusmodel,Ke20eccv-gsnet,Kundu18cvpr-3dRCNN} and references therein for details of this line of research.

{\bf Contribution}. Our first contribution, as described in the previous paragraphs, is to \emph{unify} five pose estimation problems under the general framework of aligning two sets of geometric primitives. While such proposition has been presented in~\cite{Briales17cvpr-registration,Olsson09pami-bnbRegistration} for point-to-point, point-to-line and point-to-plane correspondences, generalizing it to a broader class of primitives such as cylinders, cones, spheres, and ellipsoids, and showing its modeling capability in category-level registration (using the idea of \sues) and pose estimation given 2D-3D correspondences has never been done. Our second contribution is to develop a simple, general, intuitive, yet effective and efficient framework to solve all five examples by simulating \emph{rigid body dynamics}. As we will detail in Section~\ref{sec:DAMP}, the general formulation~\eqref{eq:primitivealignment} allows us to model $\calY$ as a \emph{fixed} rigid body and $\calX$ as a \emph{moving} rigid body with $\MT$ representing the relative pose of $\calX$ \wrt~$\calY$. We then place \emph{virtual} springs between points in $X_i$ and $Y_i$ that attain the shortest distance $\dist(\MT\transform X_i, Y_i)$ given $\MT$. The virtual springs naturally exert forces under which $\calX$ is pulled towards $\calY$ with motion governed by Newton-Euler rigid body dynamics, and moreover, the \emph{potential energy} of the dynamical system coincides with the objective function of problem~\eqref{eq:primitivealignment}. By assuming $\calX$ moves in an environment with constant damping, the dynamical system will eventually arrive at an \emph{equilibrium} point, from which a solution to problem~\eqref{eq:primitivealignment} can be obtained. Our construction of such a dynamical system is inspired by recent work on physics-based registration~\cite{Golyanik16CVPR-gravitationalRegistration,Golyanik19ICCV-acceleratedGravitational,Jauer18PAMI-physicsBasedRegistration}, but goes much beyond them in showing that simulating dynamics can solve broader and more challenging pose estimation problems other than just point cloud registration. We name our approach \emph{DynAMical Pose estimation} (\nameshort), which we hope to stimulate the connection between computer vision and dynamical systems. We evaluate \nameshort~on both simulated and real datasets (Section~\ref{sec:experiments}) and demonstrate (i) \nameshort always returns the \emph{globally optimal} solution to Examples~\ref{ex:pointcloudregistration}-\ref{ex:categoryregistration} with 3D-3D correspondences; (ii) although \nameshort converges to suboptimal solutions given 2D-3D correspondences (Examples~\ref{ex:absolutepose}-\ref{ex:categoryabsolutepose}) with very low probability ($<1\%$), using a simple scheme for escaping local minima, \nameshort almost always succeeds. Our last contribution (Section~\ref{sec:convergence}) is to (partially) demystify the surprisingly good empirical performance of \nameshort and prove a nontrivial global convergence result in the case of point cloud registration, by charactering the local stability of equilibrium points. Extending the analysis to other examples remains open.
\section{Geometry and Dynamics}
\label{sec:preliminary}

In this section, we present two key results underpinning the \nameshort algorithm. One is geometric and concerns computing the shortest distance between two geometric primitives, the other is dynamical and concerns simulating Newton-Euler dynamics of an $N$-primitive system.

\subsection{Geometry}
In view of Black-Box Optimization~\cite{Nesterov18book-convexOptimization}, the question that needs to be answered before solving problem~\eqref{eq:primitivealignment} is to \emph{evaluate} the cost function at a given $\MT \in \SEthree$, because the $\dist(X,Y)$ function is itself a minimization. Although in the simplest case of point cloud registration, $\dist(X,Y) = \norm{\vxx-\vy}$ can be written analytically, the following theorem states that in general $\dist(\cdot,\cdot)$ may require nontrivial computation. 

\begin{theorem}[Shortest Distance Pair]
\label{thm:shortestdistance}
Let $X$ and $Y$ be two primitives of types~\ref{item:point}-\ref{item:ellipsoid}, define 
$\pair{X,Y}$ as the set of points that attain the shortest distance between $X$ and $Y$,~\ie,
\bea
\pair{X,Y}\triangleq\argmin_{(\vxx,\vy) \in X \times Y} \norm{\vxx - \vy}.\label{eq:mindist}
\eea
In the following cases, $\pair{X,Y}$ (and hence $\dist(X,Y)$) can be computed either analytically or numerically.
\begin{enumerate}\itemsep-0.2em
\item \label{thm:point-point} Point-Point (\pp), $X=P(\vxx)$, $Y=P(\vy)$:
\bea
\pair{X,Y} = \{(\vxx,\vy)\}.
\eea

\item \label{thm:point-line} Point-Line (\pl), $X=P(\vxx)$, $Y=L(\vy,\vv)$:
\bea
\pair{X,Y} = \{ (\vxx,\vy+\alpha\vv): \alpha = \vv\tran(\vxx - \vy)\},
\eea
where $\vy + \alpha\vv$ is the projection of $\vxx$ onto the line.

\item \label{thm:point-plane} Point-Plane (\ph), $X=P(\vxx)$, $Y=H(\vy,\vn)$:
\bea
\pair{X,Y} = \{ (\vxx,\vxx+\alpha\vn): \alpha = \vn\tran(\vy - \vxx)  \},
\eea
where $\vxx+\alpha\vn$ is the projection of $\vxx$ onto the plane.

\item \label{thm:point-sphere} Point-Sphere (\ps), $X = P(\vxx)$, $Y=S(\vy,r)$:
\bea
\hspace{-2mm} \pair{X,Y}\!=\!\begin{cases}
 \{(\vxx,\vz): \vz \in S(\vy,r) \} &\!\!\!\! \text{if } \vxx = \vy \\
 \{(\vxx, \vy + r \vv): \vv = \frac{\vxx - \vy}{\norm{\vxx - \vy}}\} &\!\!\!\! \text{otherwise} 
\end{cases}\!,\!\!\!\!
\eea
where if $\vxx$ coincides with the center of the sphere, then the entire sphere achieves the shortest distance, while otherwise $\vy + r\vv$, the projection of $\vxx$ onto the sphere, achieves the shortest distance.

\item \label{thm:point-cylinder} Point-Cylinder (\pc), $X = P(\vxx)$, $Y=C(\vy,\vv,r)$:
\bea
\pair{X,Y}\!=\!\begin{cases}
\{(\vxx,\hatvy+r\vu): \vu \in \usphere{2},\vu \perp \vv \} & \!\!\!\!\text{if } \vxx = \hatvy \\
\{(\vxx,\hatvy+r \frac{\vxx - \hatvy}{\norm{\vxx - \hatvy}})  \} &\!\!\!\!\text{otherwise}
\end{cases}\!\!,\!\!\!\!
\eea
where $\hatvy\triangleq \vy + \alpha \vv,\alpha = \vv\tran(\vxx-\vy)$, is the projection of $\vxx$ onto the central axis $L(\vy,\vv)$. If $\vxx$ lies on the central axis, then any point on the circle that passes through $\vxx$ and is orthogonal to $\vv$ achieves the shortest distance, otherwise, the projection of $\vxx - \hatvy$ onto the cylinder achieves the shortest distance.

\item \label{thm:point-cone} Point-Cone (\pk), $X = P(\vxx)$, $Y=K(\vy,\vv,\theta)$:
\bea
\hspace{-8mm} \pair{X,Y}\!\!=\!\!\begin{cases}
\!\{ (\vxx,\vy) \} &\!\!\!\hspace{-20mm}\text{if } \vv\tran\vxx_y \leq - \norm{\vxx_y}\sin\theta\\
\!\{(\vxx,\vy + \norm{\vxx_y}\cos\theta \vu) :\substack{\vu \in \usphere{2}, \\ \vu\tran\vv = \cos\theta} \} &\!\!\!\text{if } \frac{\vxx_y}{\norm{\vxx_y}} = \vv \\
\!\{(\vxx,\vy+\alpha\vw): \alpha = \vw\tran\vxx_y \} &\!\!\!\text{otherwise}
\end{cases}\!\!,\!\!\!\!
\eea
where $\vxx_y \triangleq \vxx - \vy$, $\vw \triangleq \MR_\theta \vv$, with $\MR_\theta \in \SOthree$ being the 3D rotation matrix of axis $\vv \times \frac{\vxx_y}{\norm{\vxx_y}}$ and angle $\theta$.\footnote{$\va \times \vb$ denotes the cross product of $\va,\vb \in \Real{3}$. Given an axis-angle representation $(\vv,\theta)$ of a 3D rotation, the rotation matrix can be computed as $\MR = \cos\theta \eye_3 + \sin\theta \hatmap{\vv} + (1-\cos\theta)\vv\vv\tran$, where $\hatmap{\vv}$ is the skew-symmetric matrix associated with $\vv$ such that $\vv \times \va \equiv \hatmap{\vv}\va$~\cite{Yang19iccv-QUASAR}.} The first condition $\vv\tran\vxx_y \leq - \norm{\vxx_y}\sin\theta$ corresponds to $\vxx$ in the dual cone of $K$ and the apex $\vy$ achieves the shortest distance. The second condition corresponds to $\vxx$ lies on the central axis and inside the cone, in which case an entire circle on the surface of the cone achieves the shortest distance. Under the last condition, a unique projection of $\vxx$ onto (an extreme ray of) the cone achieves the shortest distance. 

\item \label{thm:point-ellipsoid} Point-Ellipsoid (\pe), $X = P(\vxx)$, $Y=E(\vy,\MA)$:
\bea
\hspace{-8mm} \pair{X,Y}\!=\!\begin{cases}
\!\{(\vxx,\vxx)\} & \!\!\!\!\text{if } \vxx \in E \\
\!\{(\vxx,(\lambda \MA + \eye)\inv\vxx_y + \vy):\substack{g(\lambda)=0,\\\lambda>0}\} & \!\!\!\! \text{otherwise}
\end{cases}\!\!,\!\!
\eea
where $\vxx_y \triangleq \vxx - \vy$, and $g(\lambda)$ is a univariate function whose expression is given in \supp.
If $\vxx$ belongs to the ellipsoid, then the shortest distance is zero. Otherwise, there is a unique point on the surface of the ellipsoid that achieves the shortest distance, obtained by finding the root of the function $g(\lambda)$.

\item \label{thm:ellipsoid-line} Ellipsoid-Line (\el), $X = E(\vxx,\MA)$, $Y=L(\vy,\vv)$:
\bea
\hspace{-8mm} \pair{X,Y}\!=\!\begin{cases}
\!\{(\vy+\alpha\vv,\vy+\alpha\vv): \alpha \in [\alpha_1,\alpha_2]\} &\!\!\!\! \text{if } \Delta \geq 0 \\
\!\{(\vzz(\lambda),\vy+\alpha(\lambda)\vv):\substack{g(\lambda) = 0,\\\lambda > 0}\}&\!\!\!\! \text{otherwise} 
\end{cases}\!\!,\!\!
\eea
where $\vy_x\triangleq \vy - \vxx$, 
and the expressions of $\Delta,\alpha_{1,2},\vz(\lambda),\alpha(\lambda),g(\lambda)$ are given in \supp.
Intuitively, the discriminant $\Delta$ decides when the line intersects with the ellipsoid. If there is nonempty intersection, then an entire line segment (determined by $\alpha_{1,2}$) achieves shortest distance zero. Otherwise, the unique shortest distance pair can be obtained by first finding the root $\lambda$ of a univariate function $g(\lambda)$ and then substituting $\lambda$ into $\vz(\lambda)$ and $\alpha(\lambda)$.
\end{enumerate}
\end{theorem}
A detailed proof of Theorem~\ref{thm:shortestdistance} is in \supp, with numerical methods for finding roots of $g(\lambda)$.

\begin{remark}[Distance] The $\dist(\cdot,\cdot)$ function defined in \eqref{eq:mindist} is inherited from convex analysis \cite{dax06LAA-distance} and is appropriate for problems in this paper. However, it can be ill-defined for, \eg, aligning a pyramid to a sphere. A potentially better distance function would be the \emph{Hausdorff distance} \cite{rockafellar09bookvariational}, but it is much more complicated to compute.
\end{remark}

\subsection{$N$-Primitive Rigid Body Dynamics}

\begin{figure}[h]
\centering%
\includegraphics[width=0.75\columnwidth]{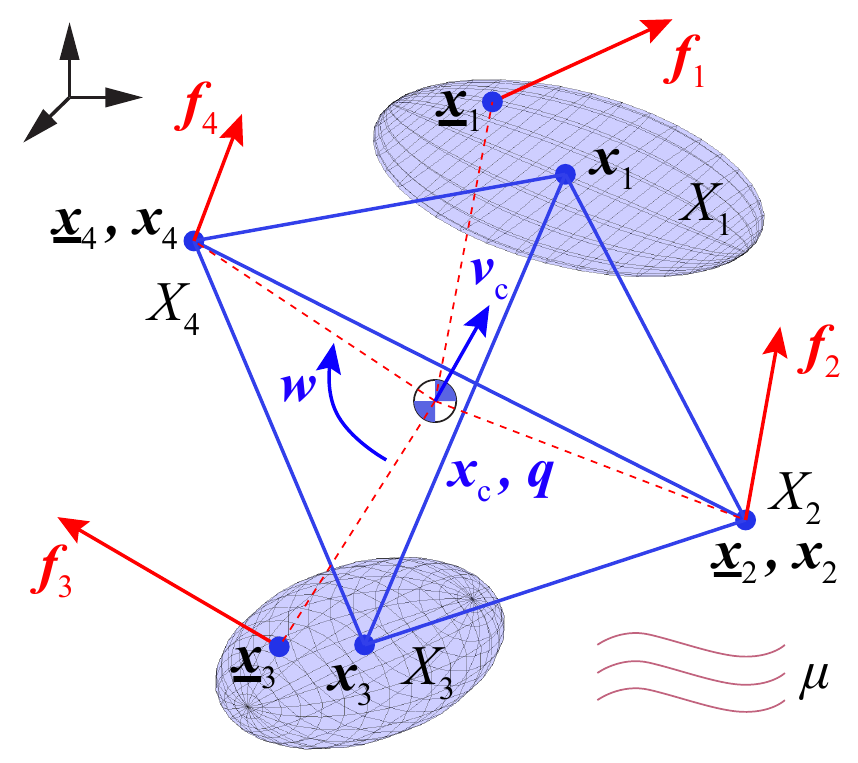}
\caption{Example of an $N$-primitive rigid-body dynamical system with $N=4$. $X_{1,3}$ are ellipsoids, $X_{2,4}$ are points. 
}
\label{fig:rigid-body-dynamics}
\vspace{-4mm}
\end{figure}

In this paper we consider a rigid body consisting of $N$ primitives $\{X_i\}_{i=1}^N$ moving in an environment with constant damping coefficient $\mu > 0$, and each primitive $X_i$ has a \emph{pointed} mass located at $\vxx_i \in \Real{3}$ \wrt a \emph{global} coordinate frame (Fig.~\ref{fig:rigid-body-dynamics}). Assume there is an external force $\vf_i \in \Real{3}$ acting on each primitive at location $\ux_i \in \Real{3}, i=1,\dots,N$. Note that we do not restrict $\vxx_i = \ux_i$,~\ie~the external force is not required to act at the location of the pointed mass. For example, when $X_i$ is an ellipsoid, $\vxx_i$ is the center of the ellipsoid, but $\ux_i$ can be any point on the surface of or inside the ellipsoid (\cf~Fig.~\ref{fig:rigid-body-dynamics}). We assume each primitive has equal mass $m_i = m,i=1,\dots,N$, such that the \emph{center of mass} of the $N$-primitive system is at $\barx \triangleq \frac{1}{N} \sum_{i=1}^N \vxx_i$ (in the global frame). The next proposition states the system of equations governing the motion of the $N$-primitive system.

\begin{proposition}[$N$-Primitive Dynamics]
\label{prop:primitivedynamics}
Let $\state(t) \triangleq [\xcm\tran,\vq\tran,\vcm\tran,\vomega\tran]\tran \in \Real{13}$ be the state space of the $N$-primitive rigid body in Fig.~\ref{fig:rigid-body-dynamics}, where $\xcm\in \Real{3}$ denotes the position of the center of mass in the global coordinate frame, $\vq \in \usphere{3}$ denotes the unit quaternion representing the rotation from the body frame to the global frame, $\vcm \in \Real{3}$ denotes the translational velocity of the center of mass, and $\vomega \in \Real{3}$ denotes the angular velocity of the rigid body \wrt the center of mass. At $t=0$, assume
\bea \label{eq:initialState}
\xcm(0) = \barx, \ \vq(0) = [0,0,0,1]\tran, \ \vcm(0) = \zero, \ \vomega(0) = \zero,
\eea
so that the body frame coincides with the global frame ($\vq(0)$ is the identity rotation). Call
$\xref{i} \triangleq \vxx_i - \barx$
the relative position of $\vxx_i$ \wrt the center of mass expressed in the body frame (a constant value \wrt time), 
 then under the external forces $\vf_i$ acted at locations $\ux_i$, expressed in global frame, the equations of motion of the dynamical system are
\bea
\hspace{-4mm} \dstate(t) = \calF(\state; \vf_i, \ux_i,\mu) =  
\begin{cases}
\dxcm = \vcm \\
\dq = \frac{1}{2} \vq \qprod \omegahomo \\
\dvcm := \acm = \frac{1}{M} \vf \\
\domega := \valpha = \MJ\inv(\vtau - \vomega \times \MJ \vomega)
\end{cases}\!\!\!\!\!\!\!,\!\! \label{eq:Nprimitivedynamics}
\eea
where $\omegahomo \triangleq [\vomega\tran,0]\tran \in \Real{4}$ is the homogenization of $\vomega$, ``$\qprod$'' denotes the quaternion product~\cite{Yang19iccv-QUASAR},
$M \triangleq N m$ is the total mass of the system, $\vf$ is the total external force
\bea
\vf = \sum_{i=1}^N \overbrace{\vf_i - \mu m (\vcm + \MR_q ( \vomega \times \xref{i} ) )}^{:=\vf_i'}, \label{eq:totalforce}
\eea
with $\MR_q \in \SOthree$ being the unique rotation matrix associated with the quaternion $\vq$, $\MJ$ is the moment of inertia $\MJ \triangleq - m \sum_{i=1}^N \hatmap{\xref{i}}^2 \in \pd{3}$
expressed in the body frame, and $\vtau$ is the total torque
\bea
\vtau = \sum_{i=1}^N \MR_q\tran (\ux_i - \xcm)\times (\MR_q\tran \vf_i' ), \label{eq:totaltorque}
\eea
in the body frame ($\MR_q\tran$ rotates vectors to body frame).
\end{proposition}
The proof of Proposition~\ref{prop:primitivedynamics} follows directly from~\cite{baraff97siggraph-rigidbody}.
\begin{remark}[Unbounded Primitives] In this paper, it suffices to consider bounded primitives (ellipsoids, points) in the $N$-primitive system. For an unbounded primitive (\eg, lines, planes), it remains open how to distribute its mass. A simple idea is to place all its mass $m_i$ at the point of contact $\ux_i$.
\end{remark}

\section{Dynamical Pose Estimation}
\label{sec:DAMP}
\subsection{Overview of \nameshort}
\label{sec:alg-overview}
The idea in \nameshort is to treat $\calX$ as the $N$-primitive rigid body in Fig.~\ref{fig:rigid-body-dynamics}, and treat $\calY$ as a set of primitives in the \emph{global frame} that stay fixed and generate external forces to $\calX$,~\ie,~each primitive $Y_i$ applies an external force $\vf_i$ on $X_i$ at location $\ux_i$ (red arrows in Fig.~\ref{fig:rigid-body-dynamics}). Although this idea is inspired by related works~\cite{Golyanik16CVPR-gravitationalRegistration,Golyanik19ICCV-acceleratedGravitational,Jauer18PAMI-physicsBasedRegistration}, our construction of the forces significantly differ from them in two aspects: (i) we place a \emph{virtual spring}, with coefficient $k$, between each pair of corresponding primitives $(X_i,Y_i)$;\footnote{Previous works~\cite{Golyanik16CVPR-gravitationalRegistration,Golyanik19ICCV-acceleratedGravitational,Jauer18PAMI-physicsBasedRegistration} use gravitational and electrostatic forces between two point clouds, under which the potential energy of the dynamical system is not equivalent to the objective function of~\eqref{eq:primitivealignment}.} (ii) the two endpoints of the virtual spring are found using Theorem~\ref{thm:shortestdistance} so that the virtual spring spans the \emph{shortest distance} between $X_i$ and $Y_i$. With this, we have the following lemma.

\begin{lemma}[Potential Energy]
\label{lemma:potentialenergy}
If the virtual spring has its two endpoints located at the shortest distance pair $\pair{\MT \transform X_i, Y_i}$ for any $\MT$, and the spring has constant coefficient $k=2$, then the cost function of problem~\eqref{eq:primitivealignment} is equal to the potential energy of the dynamical system.
\end{lemma}

We now state the \nameshort algorithm (Algorithm~\ref{alg:damp}). The input to \nameshort is two sets of geometric primitives as in problem~\eqref{eq:primitivealignment}. In particular, we require the $(X_i,Y_i)$ pair to be one of the seven types listed in Theorem~\ref{thm:shortestdistance}, which encapsulate Examples~\ref{ex:pointcloudregistration}-\ref{ex:categoryabsolutepose}. \nameshort starts by computing the center of mass $\barx$, the relative positions $\xref{i}$, and the moment of inertia $\MJ$ (line~\ref{line:cmandJ}) using the location of the pointed mass $\vxx_i$ of each primitive in $\calX$ (since $X_i$ is either a point or an ellipsoid among Examples~\ref{ex:pointcloudregistration}-\ref{ex:categoryabsolutepose}, $\vxx_i$ is well defined as in Fig.~\ref{fig:rigid-body-dynamics}). Then \nameshort computes the Cholesky factorization of $\MJ$ and stores the lower-triangular Cholesky factor $\ML$ (line~\ref{line:choleskyJ}), which will later be used to compute the angular acceleration $\accang$ in eq.~\eqref{eq:Nprimitivedynamics}.\footnote{One can also invert $\MJ$ directly since $\MJ$ is a $3\times 3$ small matrix.} In line~\ref{line:initialization}, the simulation is initialized at $\state_0$ as in~\eqref{eq:initialState}, which basically states that $\calX$ starts at rest without any initial speed. At each iteration of the main loop, \nameshort first computes a shortest distance pair $(\ux_i,\uy_i)$ between the fixed $Y_i$ and the $X_i$ at current state $\state$, denoted as $X_i(\state)$ (line~\ref{line:shortestdistance}). 
With the shortest distance pair $(\ux_i,\uy_i)$, \nameshort spawns an instantaneous virtual spring between $X_i$ and $Y_i$ with endpoints at $\ux_i$ and $\uy_i$, leading to a virtual spring force $\vf_i = k (\uy_i - \ux_i)$ (line~\ref{line:springforce}). Then \nameshort computes the time derivative of the state $\dstate$ using eqs.~\eqref{eq:Nprimitivedynamics}-\eqref{eq:totaltorque} (line~\ref{line:computedstate}). If $\norm{\dstate}$ is smaller than the predefined threshold $\varepsilon$, then the dynamical system has reached an equilibrium point and the simulation stops (line~\ref{line:breakequilibrium}). Otherwise, \nameshort updates the state of the dynamical system, with proper \emph{renormalization} on $\vq$ to ensure a valid 3D rotation (line~\ref{line:updatestate}). The initial pose of $\calX$ is $(\barx,\eye_3)$, and the final pose of $\calX$ is $(\xcm,\MR_q)$, therefore, \nameshort returns the alignment $\MT$ that transforms $\calX$ from the initial state to the final state (line~\ref{line:returnsolution}): $\MR = \MR_q,\ \vt = \xcm - \MR_q \barx$.

\setlength{\textfloatsep}{5pt}%
\begin{algorithm}[t]
\DontPrintSemicolon
\SetAlgoLined
\textbf{Input:} Primitives $\calX = \{ X_i\}_{i=1}^N$ and $\calY = \{ Y_i\}_{i=1}^N$; damping $\mu > 0$ (default: $\mu = 2$); mass $m>0$ (default: $m=1$); spring coefficient $k>0$ (default: $k=2$); boolean: \escape (default: \false); number of trials $T_{\max} > 0$ (default: $T_{\max} =5$); equilibrium threshold $\varepsilon > 0$ (default: $\varepsilon = 10^{-6}$); step size $dt > 0$ (default: $dt = 0.3$); maximum number of steps $K_{\max}$ (default: $K_{\max}=10^3$)  \\
\textbf{Output:} an estimate $\MT \in \SEthree$ to problem~\eqref{eq:primitivealignment} \\

\grayout{\% Compute $N$-primitive quantities of $\calX$} \\
$\displaystyle \barx = \frac{\sum_{i=1}^N \vxx_i}{N},\xref{i} = \vxx_i - \barx,\MJ = -m \sum_{i=1}^N \hatmap{\xref{i}}^2$ \label{line:cmandJ}\\
$\MJ = \ML \ML\tran$, with $\ML$ lower triangular \label{line:choleskyJ} \\
\grayout{\% Initialization } \\
$\state = \state_0$ as in eq.~\eqref{eq:initialState} \label{line:initialization} \\
\grayout{\% Simulate dynamics} \\
\lIf{\escape}{
	$T=0$, $\calS = \emptyset$, $\calC = \emptyset$
}
\For{$j=1,\dots,K_{\max}$}{
	\grayout{\% Compute shortest distance pair (Theorem~\ref{thm:shortestdistance})} \\
	$(\ux_i, \uy_i) \in \pair{X_i(\state), Y_i}, i=1,\dots,N$ \label{line:shortestdistance} \\
	$\vf_i = k (\uy_i - \ux_i), i=1,\dots,N$ \label{line:springforce} \\ 
	\grayout{\% Compute derivative of state (Proposition~\ref{prop:primitivedynamics})}\\
	$\dstate = \calF(\state;\vf_i,\ux_i,\mu)$, with $\MJ = \ML \ML\tran$ in~\eqref{eq:Nprimitivedynamics} \label{line:computedstate}\\
	\grayout{\% Check equilibrium} \\
	\If{$\norm{\dstate} < \varepsilon$ }{ 
		\uIf{\escape and $T \leq T_{\max}$ }{
			$\calS = \calS \cup \state$, $\calC = \calC \cup \frac{k}{2}\sum_{i=1}^N \|\uy_i - \ux_i\|^2$\\
			$\dstate \sim \calN(\zero,\eye_{13})$ \grayout{\% Random perturbation}\label{line:perturb}\\
			$T = T + 1$
		}\Else{
			{\bf break} \label{line:breakequilibrium}
		}	
	}
	\grayout{\% Update state with quaternion correction} \\
	$\state = \state + dt \cdot \dstate$, $\quad \state(\vq) \leftarrow \frac{\state(\vq)}{\norm{\state(\vq)}}$ \label{line:updatestate}
}
\lIf{\escape}{$\state = \calS ( \argmin \calC )$} \label{line:minpotentialenergy}

{\bf Return: } $\MT = (\MR_q, \xcm - \MR_q \barx)$ \label{line:returnsolution}
\caption{\nameshort \label{alg:damp}}
\end{algorithm}

{\bf Escape local minima}. The \nameshort framework allows a simple scheme for escaping suboptimal solutions. If the boolean flag \escape is \true, then each time the system reaches an equilibrium point, \nameshort computes the potential energy of the system (which is the cost function of~\eqref{eq:primitivealignment} by Lemma~\ref{lemma:potentialenergy}), stores the energy and state in $\calC$, $\calS$, and \emph{randomly perturbs} the derivative of the state (imagine a virtual ``hammering'' on $\calX$, line~\ref{line:perturb}). After executing the \escape scheme for a number of $T_{\max}$ trials, \nameshort uses the state with \emph{minimum} potential energy (line~\ref{line:minpotentialenergy}) to compute the final solution $\MT$.


\subsection{Global Convergence: Point Cloud Registration}
\label{sec:convergence}
Due to the external damping $\mu$, \nameshort is guaranteed to converge to an equilibrium point with $\dstate = \zero$, a result that is well-known from Lyapunov theory~\cite{Slotine91book-nonlinearcontrol}. However, the system~\eqref{eq:Nprimitivedynamics} may have many (even infinite) equilibrium points. Therefore, a natural question is: \emph{Does \nameshort converge to an equilibrium point that minimizes the potential energy of the system?} If the answer is affirmative, then by Lemma~\ref{lemma:potentialenergy}, we can guarantee that \nameshort finds the global minimizer of problem~\eqref{eq:primitivealignment}. The next theorem establishes the global convergence of \nameshort for point cloud registration.

\begin{theorem}[Global Convergence]
\label{thm:globalconvergence}
In problem~\eqref{eq:primitivealignment}, let $\calX$ and $\calY$ be two sets of 3D points under \emph{generic configuration}.
\begin{enumerate}[label=(\roman*)]\itemsep-0.2em
	\item \label{item:foursolution} The system~\eqref{eq:Nprimitivedynamics} has four equilibrium points ($\dstate = \zero$);
	\item \label{item:optimalsolution} One of the (optimal) equilibrium point minimizes the potential energy;
	\item \label{item:threebypi} Three other \emph{spurious} equilibrium points differ from the optimal equilibrium point by a rotation of $\pi$;
	\item \label{item:localunstable} The spurious equilibrium points are \emph{locally unstable}.
\end{enumerate}
Therefore, \nameshort (Algorithm~\ref{alg:damp} with \escape= \false) is guaranteed to converge to the optimal equilibrium point.
\end{theorem}
The proof of Theorem~\ref{thm:globalconvergence} is algebraically involved and is presented in the \supp. The condition ``generic configuration'' helps remove pathological cases such as when the 3D points are collinear and coplanar (examples given in \supp). 

\section{Experiments}
\label{sec:experiments}
We first show that \nameshort always converges to the optimal solution given 3D-3D correspondences (Section~\ref{sec:exp-3d-3d}), then we show the \escape~scheme helps escape suboptimal solutions given 2D-3D correspondences (Section~\ref{sec:exp-2d-3d}).

\subsection{3D-3D: Empirical Global Convergence}
\label{sec:exp-3d-3d}
{\bf Point Cloud Registration}. We randomly sample $N=100$ 3D points from $\calN(\zero,\eye_3)$ to be $\calX$, then generate $\calY$ by applying a random rigid transformation $(\MR,\vt)$ to $\calX$, followed by adding Gaussian noise $\calN(\zero,0.01^2\eye_3)$. We run \nameshort without $\escape$, and compare its estimated pose \wrt the groundtruth pose, as well as the \emph{optimal} pose returned by Horn's method~\cite{Horn87josa} (label: \svd). Table~\ref{table:pcrresult} shows the rotation ($\errR$) and translation $(\errt)$ estimation errors of \nameshort and \svd \wrt groundtruth, as well as the difference between \nameshort and \svd estimates ($\diffR$ and $\difft$), under $1000$ Monte Carlo runs. The statistics show that (i) \nameshort always converges to the globally optimal solution ($\diffR,\difft$ are numerically zero), empirically proving the correctness of Theorem~\ref{thm:globalconvergence}; (ii) \nameshort returns accurate pose estimations. On average, \nameshort converges to the optimal equilibrium point in $27$ iterations ($\norm{\dstate}<10^{-6}$), and runs in $6.3$ milliseconds. Although \nameshort is slower than \svd in 3D, it opens up a new method to perform high-dimensional point cloud registration by using \emph{geometric algebra}~\cite{Doran03book-GA} to simulate rigid body dynamics~\cite{Bosch20TOG-nDrigidbody}, when \svd becomes expensive. We also use the \bunny dataset for point cloud registration and \nameshort always returns the correct solution, shown in Fig.~\ref{fig:all-apps}(a).
\begin{table}[h]
\vspace{-3mm}
\adjustbox{max width=\columnwidth}{%
\centering
\begin{tabular}{ccc}
\hline
 & \nameshort & \svd~\cite{Horn87josa} \\
\hline 
$\errR\ (^\circ)$ & $(0.065/0.011/0.188)$ & $(0.065/0.011/0.188)$ \\
$\errt\ (\mathrm{m})$ & $(1.6\mathrm{e}{-3}/1.4\mathrm{e}{-4}/4.3\mathrm{e}{-3})$ & $(1.6\mathrm{e}{-3}/1.4\mathrm{e}{-4}/4.3\mathrm{e}{-3})$ \\
$\diffR\ (^\circ)$ & \multicolumn{2}{c}{$(2.9\mathrm{e}{-5}/0/5.1\mathrm{e}{-5})$} \\
$\difft\ (\mathrm{m})$ & \multicolumn{2}{c}{$(2.3\mathrm{e}{-7}/6.1\mathrm{e}{-9}/6.9\mathrm{e}{-7})$}  \\
\hline 
\end{tabular}}
\vspace{-3mm}
\caption{Point cloud registration: \nameshort converges to the globally optimal solution. Errors in $(\mathrm{mean}/\min/\max)$.}
\label{table:pcrresult}
\vspace{-2mm}
\end{table}

{\bf Primitive Registration}. In order to test \nameshort's performance on primitive registration and verify its global convergence, we follow the test setup in~\cite{Briales17cvpr-registration} using random registration problems with point-to-point, point-to-line and point-to-plane correspondences, and compare \nameshort with the state-of-the-art \emph{certifiably optimal} solver in~\cite{Briales17cvpr-registration} based on semidefinite relaxation (label: \sdr). In particular, we randomly sample 50 points, 50 lines and 50 planes (150 primitives in total) within a scene with radius 10, randomly sample a point on each primitive, and transform the sampled points by a random $(\MR,\vt)$, followed by adding Gaussian noise $\calN(\zero,\sigma^2 \eye_3)$. We increase the noise level $\sigma$ from $0.01$ to $2$, and perform 1000 Monte Carlo runs at each noise level. Fig.~\ref{fig:mesh-R-time} boxplots the rotation estimation error and runtime of \nameshort and \sdr (SDP solved by SeDuMi~\cite{Sturm99-sedumi} with CVX interface~\cite{CVXwebsite}). We observe that (i) \nameshort always returns the same solution as \sdr, which is certified to be the globally optimal solution (\supp~plots the relative duality gap of \sdr is always zero); (ii) \nameshort is about 10 times faster than \sdr, despite being implemented in Matlab using for loops. The translation error looks similar as rotation error and is shown in \supp. This experiment shows that the same global convergence Theorem~\ref{thm:globalconvergence} is very likely to hold in the case of general primitive registration with line and plane correspondences. In fact, \supp~also performs the same set of experiments using the robot primitive model in Fig.~\ref{fig:all-apps}(b) with spheres, cylinders and cones, demonstrating that \nameshort also \emph{always} converges to an accurate (most likely optimal) pose estimate (note that we cannot claim global optimality because there is no guaranteed globally optimal solver, such as \sdr~\cite{Briales17cvpr-registration}, in that case to verify \nameshort).

\newcommand{\mpwsingletwo}{4.5cm}
\begin{figure}[h]
\vspace{-2mm}
\begin{minipage}{\textwidth}
\begin{tabular}{cc}%
\hspace{-6mm}\begin{minipage}{\mpwsingletwo}%
\centering%
\includegraphics[width=\columnwidth]{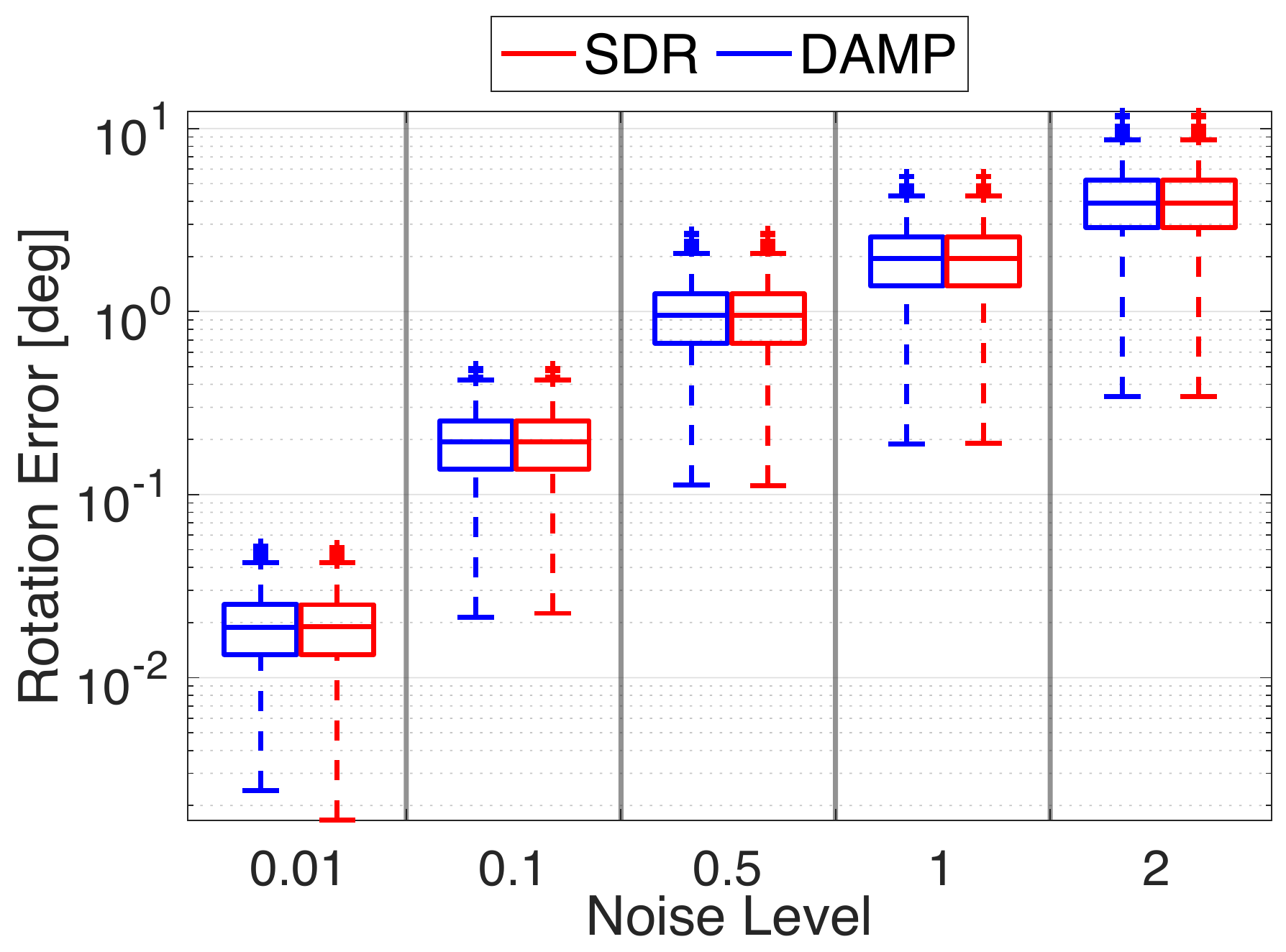}
\end{minipage}
&
\hspace{-5mm} \begin{minipage}{\mpwsingletwo}%
\centering%
\includegraphics[width=\columnwidth]{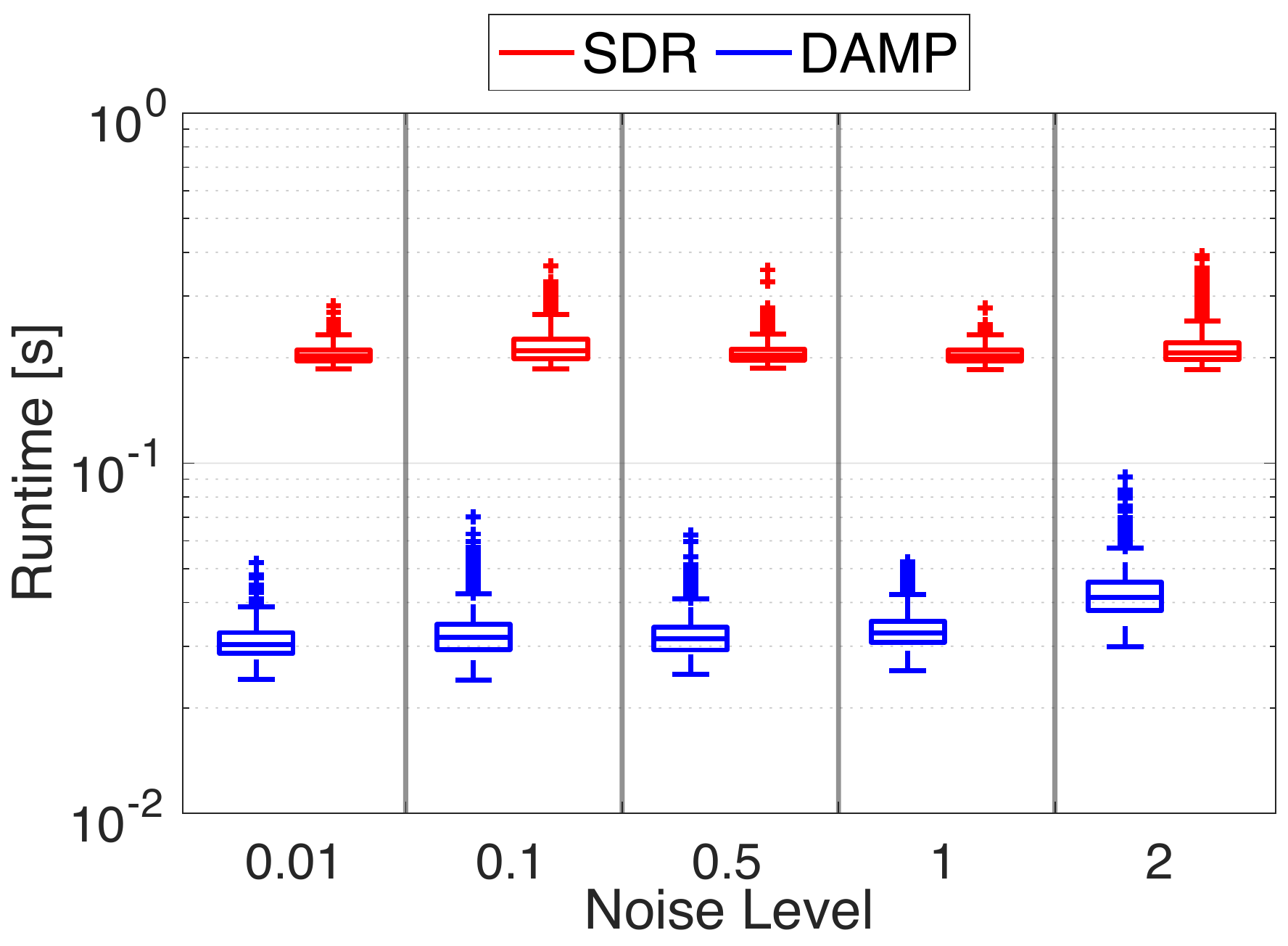}
\end{minipage}
\end{tabular}
\end{minipage}
\vspace{-3mm}
\caption{Rotation error and runtime of \nameshort compared with \sdr~\cite{Briales17cvpr-registration} on random primitive registration with increasing noise levels. \nameshort always converges to the globally optimal solution while being 10 times faster.}
\label{fig:mesh-R-time}
\vspace{-4mm}
\end{figure}

{\bf Category Registration}. We use three categories, \emph{aeroplane}, \emph{car}, and \emph{chair}, from the \pascal dataset~\cite{Xiang2014WACV-PASCAL+} to test \nameshort for category registration. In particular, given a list of $K$ instances in a category, where each instance has $N$ semantic keypoints $\calB_k \in \Real{3 \times N},k=1,\dots,K$. We first build a category model of the $K$ instances into $N$ \sues~(see \supp) and use it as $\calY$ in problem~\eqref{eq:primitivealignment}. Then we randomly generate an unknown instance of this category by following the active shape model~\cite{Zhou17pami-shapeEstimationConvex,Yang20cvpr-shapeStar},~\ie~$\calS = \sum_{k=1}^K c_k \calB_k$ with $c_k \geq 0, \sum_{k=1}^K c_k = 1$. After this, we apply a random transformation $(\MR,\vt)$ to $\calS$ to obtain $\calX$ in problem~\eqref{eq:primitivealignment}. We have $N=8,K=8$ for aeroplane, $N=12,K=9$ for car, and $N=10,K=8$ for chair. For each category, we perform 1000 Monte Carlo runs and Fig.~\ref{fig:cr-R-t} summarizes the rotation and translation estimation errors. We can see that \nameshort returns accurate rotation and translation estimates for all 1000 Monte Carlo runs of each category. Because a globally optimal solver is not available for the case of registering a point cloud to a set of ellipsoids, we cannot claim the global convergence of \nameshort, although the results highly suggest the global convergence. An example of registering the chair category is shown in Fig.~\ref{fig:all-apps}(c).

\renewcommand{\mpwsingletwo}{4cm}
\begin{figure}[h]
\vspace{-3mm}
\begin{minipage}{\textwidth}
\begin{tabular}{cc}%
\hspace{-3mm} \begin{minipage}{\mpwsingletwo}%
\centering%
\includegraphics[width=0.9\columnwidth]{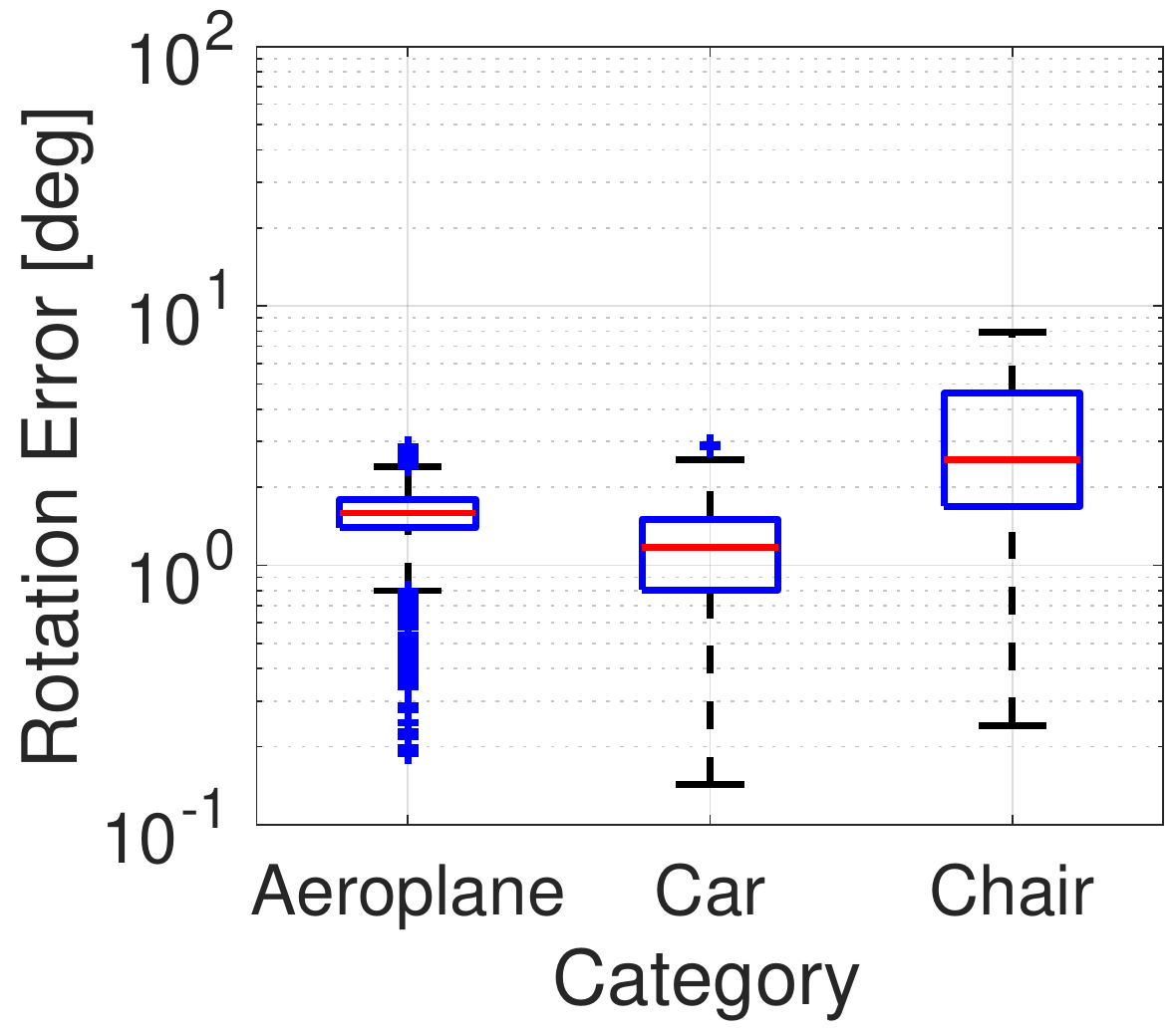}
\end{minipage}
&
\hspace{-3mm} \begin{minipage}{\mpwsingletwo}%
\centering%
\includegraphics[width=0.9\columnwidth]{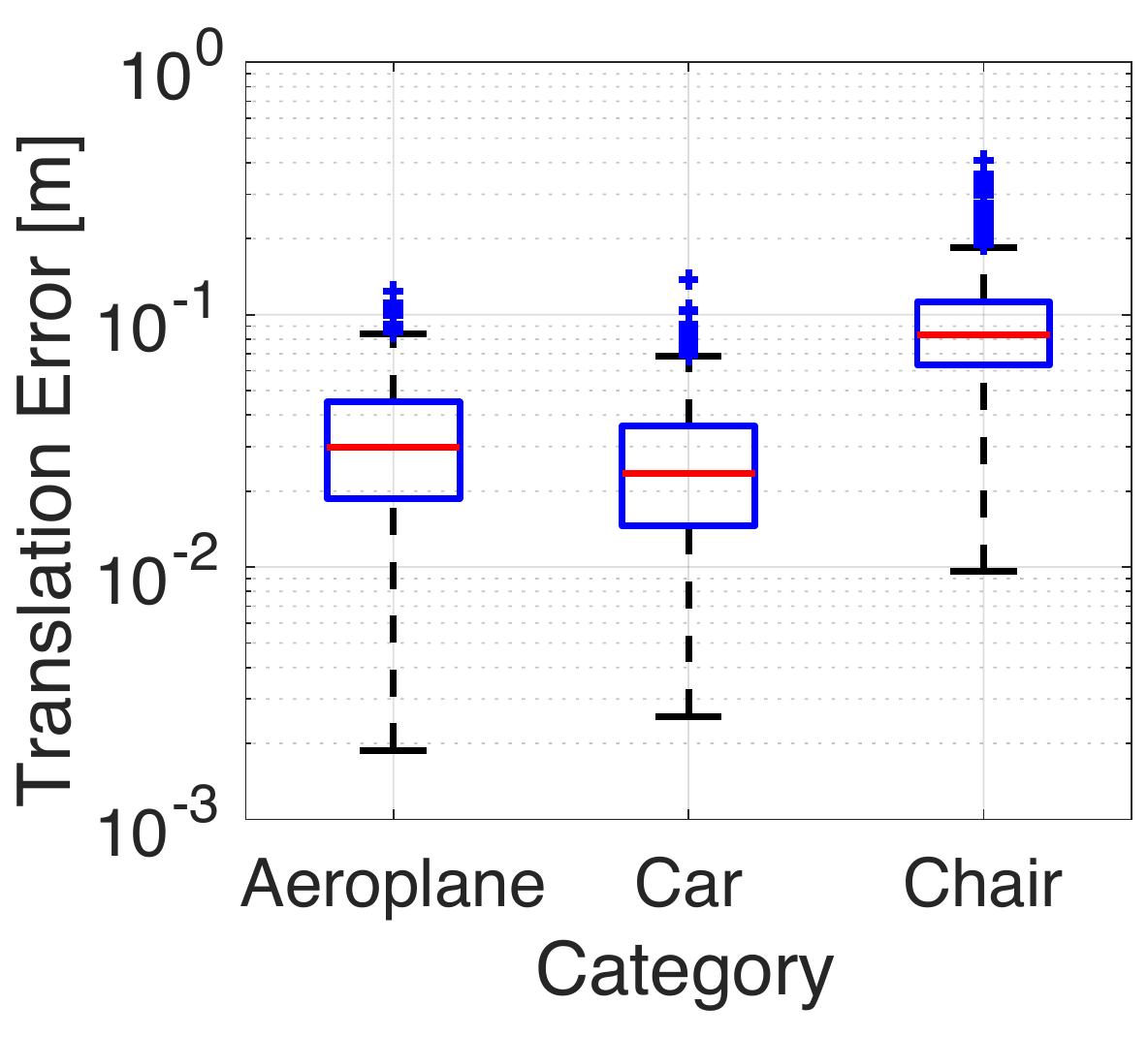}
\end{minipage}
\end{tabular}
\end{minipage}
\vspace{-3mm}
\caption{Rotation and translation estimation error of \nameshort on category registration using the aeroplane, car, and chair categories from~\pascal dataset~\cite{Xiang2014WACV-PASCAL+}.}
\label{fig:cr-R-t}
\vspace{-4mm}
\end{figure}

\subsection{2D-3D: Escape Local Minima}
\label{sec:exp-2d-3d}
{\bf Absolute Pose Estimation}. We follow the protocol in~\cite{Kneip2014ECCV-UPnP} for absolute pose estimation. We first generate $N$ groundtruth 3D points within the $[-2,2]\times[-2,2]\times[4,8]$ box inside the camera frame, then project the 3D points onto the image plane and add random Gaussian noise $\calN(\zero,0.01^2 \eye_2)$ to the 2D projections. $N$ bearing vectors are then formed from the 2D projections to be the set $\calY$ in problem~\eqref{eq:primitivealignment}. We apply a random $(\MR,\vt)$ to the groundtruth 3D points to convert them into the world frame as the set $\calX$ in problem~\eqref{eq:primitivealignment}. We apply \nameshort to solve 1000 Monte Carlo runs of this problem for $N=50,100,200$, with both $\escape = \false$ and $\escape = \true$ ($T_{\max} = 5$). Table~\ref{table:apesuccess} shows the success rate of \nameshort, where we say a pose estimation is successful if rotation error is below $5^\circ$ and translation error is below $0.5$. One can see that, (i) even without the \escape~scheme, \nameshort already has a very high success rate and it only failed twice when $N=100$; (ii) with the \escape~scheme, \nameshort achieves a $100\%$ success rate. This experiment indicates that the special configuration of the bearing vectors (\ie,~they form a ``cone'' pointed at the camera center) is more challenging for \nameshort to converge. 
We also apply \nameshort to satellite pose estimation from 2D landmarks detected by a neural network~\cite{Chen19ICCVW-satellitePoseEstimation} using the \speed dataset~\cite{Sharma19arXiv-SPEED} and a successful example is provided in Fig.~\ref{fig:all-apps}(d).

\begin{table}[h]
\vspace{-2mm}
\adjustbox{max width=\columnwidth}{%
\centering
\begin{tabular}{ccccccc}
\hline
$N$ & \multicolumn{2}{c}{$50$} & \multicolumn{2}{c}{$100$} & \multicolumn{2}{c}{$200$} \\
\hline
Escape & \false & \true & \false & \true & \false & \true \\
\hline
Success (\%) & $100$ & $100$ & $99.8$ & $100$ & $100$ & $100$ \\
\hline
\end{tabular}}
\vspace{-3mm}
\caption{Success rate of \nameshort on absolute pose estimation with \escape~flag set to \true and \false.}
\label{table:apesuccess}
\vspace{-3mm}
\end{table} 
\begin{figure}[h]
\vspace{-3mm}
\centering
\includegraphics[width=\columnwidth]{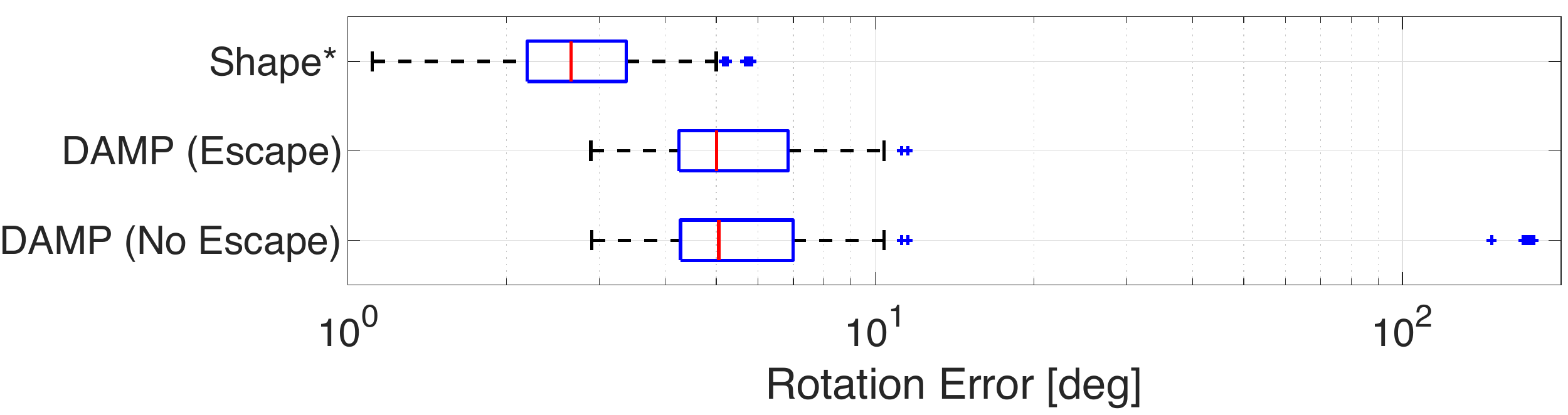}
\vspace{-7mm}
\caption{Rotation estimation error of \nameshort (both with and without~\escape) and~\shapestar on \fgthreedcar~\cite{Lin14eccv-modelFitting}.}
\label{fig:fg3dcar-rot-err}
\vspace{-3mm}
\end{figure}

{\bf Category APE}. 
We test \nameshort on~\fgthreedcar~\cite{Lin14eccv-modelFitting} for category APE, which contains 300 images of cars each with $N=256$ 2D landmark detections. \nameshort performs pose estimation by aligning the category model of \sues~(\cf~Fig.~\ref{fig:all-apps}(e)) to the set of bearing vectors. Fig.~\ref{fig:fg3dcar-rot-err} compares the rotation estimation error of \nameshort with~\shapestar~\cite{Yang20cvpr-shapeStar}, a state-of-the-art certifiably optimal solver for joint shape and pose estimation from 2D landmarks. We can see that \nameshort without \escape~fails on 6 out of the 300 images, but \nameshort with \escape succeeds on all 300 images, and return rotation estimates that are similar to \shapestar (note that the difference is due to \shapestar using a weak perspective camera model). We do notice that this is a challenging case for \nameshort because it takes more than 1000 iterations to converge, and the average runtime is 20 seconds. However, \nameshort is still faster than \shapestar (about 1 minute runtime), and we believe there is significant room for speedup by using parallelization~\cite{Jauer18PAMI-physicsBasedRegistration,Ali20arxiv-fastgravitational}.

\section{Conclusion}
\label{sec:conclusion}
We proposed \nameshort, the first general meta-algorithm for solving five pose estimation problems by simulating rigid body dynamics. We demonstrated surprising global convergence of \nameshort: it always converges given 3D-3D correspondences, and effectively escapes suboptimal solutions given 2D-3D correspondences. We proved a global convergence result in the case of point cloud registration.

Future work can be done to (i) extend the global convergence to general primitive registration; (ii) explore GPU parallelization~\cite{Ali20arxiv-fastgravitational} to enable a fast implementation; (iii) generalize \nameshort to high-dimensional registration for applications such as unsupervised language translation \cite{Conneau18iclr-word,Artetxe18acl-robust}. Geometric algebra (GA)~\cite{Doran03book-GA} can describe rigid body dynamics in any dimension, but computational challenges remain in high-dimensional GA and deserve further investigation.

\clearpage

\begin{center}
\emph{\bf {\Large Supplementary Material} }
\end{center}

\renewcommand{\thetheorem}{A\arabic{theorem}}
\renewcommand{\thesection}{A\arabic{section}}
\renewcommand{\theequation}{A\arabic{equation}}
\renewcommand{\thefigure}{A\arabic{figure}}

\setcounter{equation}{0}
\setcounter{section}{0}
\setcounter{theorem}{0}
\setcounter{figure}{0}
\setcounter{table}{0}

\section{Semantic Uncertainty Ellipsoid}
\label{app:sec:semantic-uncertainty-ellipsoid}
\begin{figure}[t]
	\begin{center}
	\begin{minipage}{\textwidth}
	\begin{tabular}{c}%
			\hspace{8mm} \begin{minipage}{6cm}%
			\centering%
			\includegraphics[width=0.9\columnwidth]{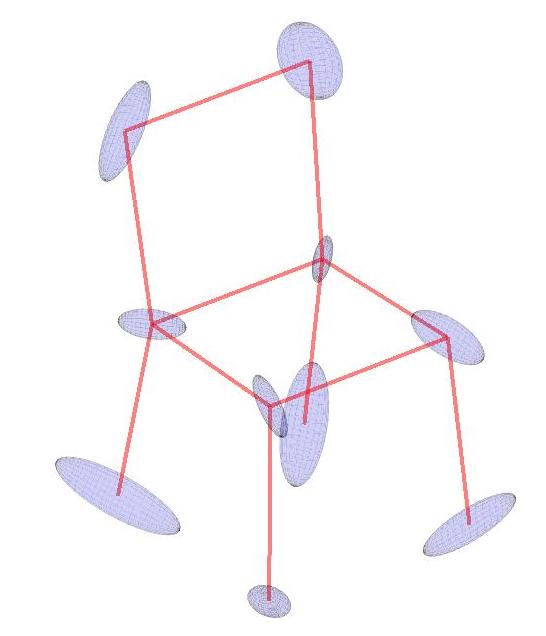} \\
			{\small (a) Chair in~\pascal~\cite{Xiang2014WACV-PASCAL+}.}
			\end{minipage}
		\\ \hspace{8mm}
			\begin{minipage}{6cm}%
			\centering%
			\includegraphics[width=0.9\columnwidth]{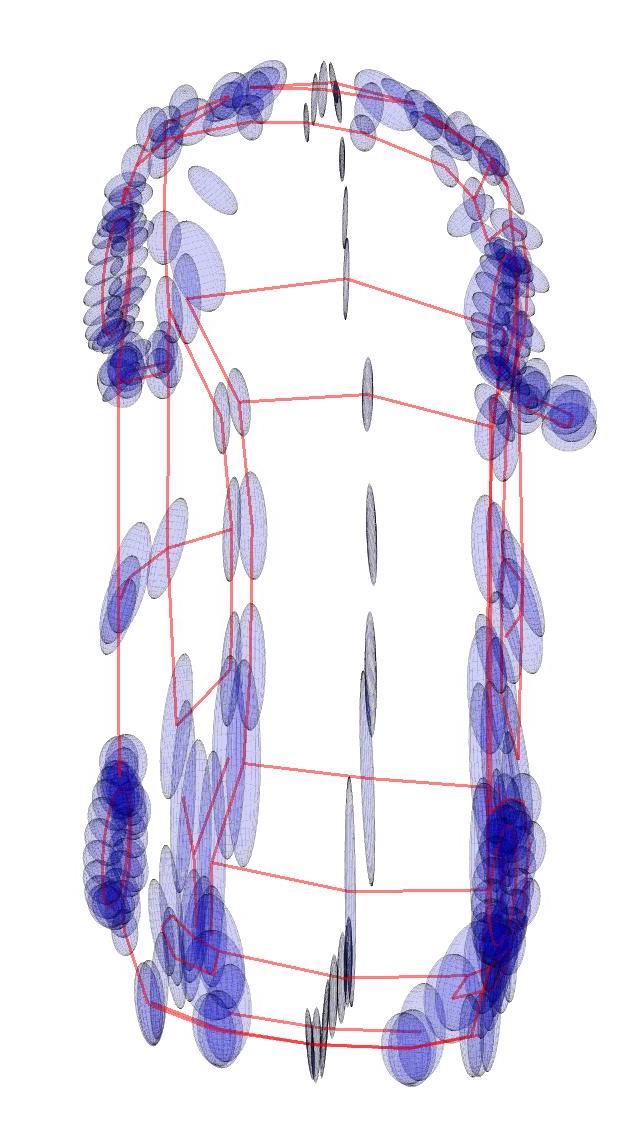} \\
			{\small (b) Car in~\fgthreedcar~\cite{Lin14eccv-modelFitting}.}
			\end{minipage}
	\end{tabular}
	\end{minipage} 
	\caption{Represent category models using a collection of semantic uncertainty ellipsoids (\sues).
	\label{fig:app-sue}} 
	\end{center}
\end{figure}

The idea of a semantic uncertainty ellipsoid (\sue) is borrowed from the \emph{error ellipsoid} that is commonly used in statistics, but we apply it to category-level pose estimation for the first time. Given a library of $K$ shapes of a category, $\calB_k,k=1,\dots,K$, where each $\calB_k \in \Real{3\times N}$ contains a list of $N$ semantic keypoints. For example, in the category of car, $\calB_k$ can be different CAD models from different car manufacturers, with annotations of certain semantic keypoints that exist for all CAD models,~\eg,~wheels, mirrors. Then we build a \sue~for the $i$-th semantic keypoint as follows. We first compute the average position of the semantic keypoint as
\bea
\vb_i = \frac{1}{K} \sum_{k=1}^K \calB_k (i),
\eea
where $\calB_k (i)$ denotes the location of the $i$-th keypoint in the $k$-th shape. We then compute the covariance matrix for the $i$-th keypoint
\bea
\MC_i = \frac{1}{K} \sum_{k=1}^K (\calB_k(i) - \vb_i) (\calB_k(i) - \vb_i)\tran.
\eea
Using $\vb_i$ and $\MC_i$, we assume that the position of the $i$-th semantic keypoint, denoted as $\vxx_i$, satisfies the following multivariate Gaussian distribution:
\bea
p(\vxx_i) = \frac{\exp\parentheses{-\frac{1}{2} (\vxx_i - \vb_i)\tran \MC_i\inv (\vxx_i - \vb_i) }}{\sqrt{(2\pi)^3 \abs{\MC_i}}},
\eea
where $\abs{\MC_i} \triangleq \det{\MC_i}$ denotes the determinant of $\MC_i$. Under this assumption, it is known that the square of the Mahalanobis distance,~\ie,~$(\vxx_i - \vb_i)\tran \MC_i \inv (\vxx_i - \vb_i)$ satisfies a chi-square distribution with three degrees of freedom:
\bea
(\vxx_i - \vb_i)\tran \MC_i \inv (\vxx_i - \vb_i) \sim \chi^2_3.
\eea
Therefore, given a confidence $\eta \in (0,1)$, we have:
\bea
\mathbb{P}\parentheses{(\vxx_i - \vb_i)\tran \MC_i \inv (\vxx_i - \vb_i) \leq \chi^2_3 (\eta)} = \eta,
\eea
where $\chi^2_3(\eta)$ corresponds to the probabilistic quantile of confidence $\eta$. This states that, with probability $\eta$, the point $\vxx_i$ lies inside the 3D ellipsoid
\bea
\frac{(\vxx_i - \vb_i)\tran \MC_i \inv (\vxx_i - \vb_i)}{\chi^2_3(\eta)} \leq 1.
\eea
We call this ellipsoid the \sue~with confidence $\eta$, and in our experiments we choose $\eta = 0.5$. Fig.~\ref{fig:app-sue} shows two examples of category models with \sues.


\section{Proof of Theorem~\ref{thm:shortestdistance}}
\label{app:sec:proof-shortest-distancepair}
\begin{proof}
Results~\ref{thm:point-point}-\ref{thm:point-cone} are basic results in 3D geometry~\cite{Briales17cvpr-registration,li19cvpr-primitiveFitting} that can be verified by inspection. We now prove~\ref{thm:point-ellipsoid} and~\ref{thm:ellipsoid-line}. The proof for~\ref{thm:point-ellipsoid} is based on~\cite{Kiseliov94LMJ-pointEllipsoid}, while the proof for~\ref{thm:ellipsoid-line} is new.

{\bf Point-Ellipsoid (\pe)}. According to the definition of the shortest distance~\eqref{eq:mindist}, the point in the ellipsoid $E(\vy,\MA)$ that attains the shortest distance to $\vxx$ is the minimizer of the following optimization:
\bea \label{eq:min-point-ellipsoid}
\min_{\vz \in \Real{3}} & \norm{\vz - \vxx}^2\\
\subject & (\vzz - \vy)\tran \MA (\vzz-\vy) \leq 1.
\eea 
Problem~\eqref{eq:min-point-ellipsoid} has a single inequality constraint and hence satisfies the \emph{linear independence constraint qualification} (LICQ)~\cite{Boyd04book}. Therefore, any solution of~\eqref{eq:min-point-ellipsoid} must satisfy the KKT conditions,~\ie~there exist $(\vzz,\lambda)$ such that:
\bea
(\vzz-\vy)\tran\MA(\vzz-\vy) - 1 \leq 0 \\
\lambda \geq 0 \\
\nabla_{\vzz} \calL \triangleq 2 (\vzz - \vxx) + 2\lambda\MA (\vzz - \vy) = \zero \\
\lambda ((\vzz-\vy)\tran\MA(\vzz-\vy) - 1) = 0
\eea
where $\calL \triangleq \norm{\vzz-\vxx}^2 + \lambda ((\vzz-\vy)\tran\MA(\vzz-\vy)-1)$ is the Lagrangian. Let $\vzz_y \triangleq \vzz - \vy, \vxx_y \triangleq \vxx - \vy$, the equations above can be written as:
\bea
\vzz_y\tran \MA \vzz_y -1 \leq 0 \label{eq:zfeasible}\\
\lambda \geq 0 \\
(\lambda\MA + \eye)\vzz_y = \vxx_y \label{eq:nablaLvanish}\\
\lambda (\vzz_y\tran\MA\vzz_y - 1) = 0 \label{eq:complementarity}
\eea
Now we can discuss two cases: (i) if $\lambda=0$, then from~\eqref{eq:nablaLvanish}, we have $\vzz_y = \vxx_y$, thus $\vzz = \vxx$ attains the global minimum $\norm{\vzz - \vxx} = 0$ (the objective function is lower bounded by $0$). In order to satisfy feasibility~\eqref{eq:zfeasible}, $\vxx_y\tran\MA\vxx_y \leq 1$ must hold and $\vxx$ has to belong to the ellipsoid; (ii) if $\lambda > 0$, then from~\eqref{eq:complementarity} we have $\vzz_y\tran\MA\vzz_y = 1$ and the optimal $\vzz$ lies on the surface of the ellipsoid. Because $\lambda > 0$ and $\MA \succ 0$, $\lambda\MA + \eye$ must be invertible and eq.~\eqref{eq:nablaLvanish} yields:
\bea
\vzz_y(\lambda) = (\lambda\MA+\eye)\inv \vxx_y,
\eea
where we use $\vzz_y(\lambda)$ to indicate $\vzz_y$ as a function of $\lambda$. Substituting this expression into $\vzz_y\tran\MA\vzz_y = 1$, we have that:
\bea
g(\lambda) \triangleq \vzz_y(\lambda)\tran \MA \vzz_y(\lambda) -1 = 0.
\eea
To see how many roots $g(\lambda)$ has in the range $\lambda > 0$, we note:
\bea
g(\lambda=0) = \vxx_y\tran \MA \vxx_y - 1 \label{eq:valuezero}\\
g(\lambda\rightarrow +\infty) = -1  \label{eq:valueinfty}
\eea
and compute the derivative of $g(\lambda)$:
\bea
g'(\lambda) = 2\vzz_y'(\lambda)\tran \MA \vzz_y(\lambda) \nonumber \\
= -2 \vzz_y(\lambda)\tran \parentheses{\MA (\lambda\MA+\eye)\inv \MA} \vzz_y(\lambda) < 0 \label{eq:negativederivative}
\eea
where $\vzz_y'(\lambda)$, the derivative of $\vzz_y(\lambda)$ \wrt $\lambda$, can be obtained by differentiating both sides of eq.~\eqref{eq:nablaLvanish} \wrt $\lambda$:
\bea
\MA \vzz_y(\lambda) + (\lambda\MA+\eye)\vzz_y'(\lambda) = 0 \Rightarrow \nonumber \\
\vzz_y'(\lambda) = - (\lambda\MA+\eye)\inv \MA \vzz_y(\lambda).
\eea
The last inequality in~\eqref{eq:negativederivative} follows from the positive definiteness of the matrix $\MA (\lambda\MA+\eye)\inv \MA$. Eqs.~\eqref{eq:valuezero}-\eqref{eq:negativederivative} show that the function $g(\lambda)$ is monotonically decreasing for $\lambda>0$. Therefore, $g(\lambda)$ has a unique root in the range $\lambda > 0$ if and only if $g(0) > 0$,~\ie~$\vxx_y\tran\MA\vxx_y > 1$. Lastly, to see the solution is indeed a minimizer, observe that the Hessian of the Lagrangian~\wrt~$\vzz$ is:
\bea
\nabla_{\vzz\vzz}\calL = 2(\lambda\MA+\eye) \succ 0,
\eea 
which is positive definite, a sufficient condition for $\vzz$ to be a global minimizer (because there is a single local minimizer, it is also global), concluding the proof of~\ref{thm:point-ellipsoid}. 

We note that the proof above also provides an efficient algorithm to numerically compute the root of $g(\lambda)=0$ and find the optimal $\vzz$, using Newton's root finding method~\cite{Nocedal99}. To do so, we initialize $\lambda_0 = 0$, and iteratively perform:
\bea
\lambda_k = \lambda_{k-1} - \frac{g(\lambda_{k-1})}{g'(\lambda_{k-1})}, \quad k=1,\dots \label{eq:newtonrootfinding}
\eea
until $g(\lambda_k) = 0$ (up to numerical accuracy). This algorithm has local quadratic convergence and typically finds the root within 20 iterations (as we will show in Section~\ref{sec:experiments}). 

{\bf Ellipsoid-Line (\el)}. First we decide if the line intersects with the ellipsoid. Since any point on the line $L(\vy,\vv)$ can be written as $\vy + \alpha \vv$ for some $\alpha \in \Real{}$, the line intersects with the ellipsoid if and only if:
\bea
(\vy + \alpha \vv - \vxx)\tran \MA (\vy + \alpha \vv - \vxx) = 1 \label{eq:linecrossellipsoid}
\eea
has real solutions. Let $\vy_x \triangleq \vy - \vxx$, eq~\eqref{eq:linecrossellipsoid} simplifies as:
\bea
(\vv\tran \MA \vv) \alpha^2 + (2\vy_x\tran\MA\vv)\alpha + (\vy_x\tran\MA\vy_x -1) = 0, \label{eq:linecrossellipsoid1}
\eea
where $\vv\tran \MA \vv > 0$ due to $\MA \succ 0$. The discriminant of the quadratic polynomial is:
\bea
\Delta = 2\sqrt{(\vy_x\tran\MA\vv)^2 - (\vv\tran \MA \vv)(\vy_x\tran\MA\vy_x -1)}.
\eea 
Therefore, eq.~\eqref{eq:linecrossellipsoid1} has two roots (counting multiplicity):
\bea
\alpha_{1,2} = \frac{-\vy_x\tran\MA\vv \pm \Delta }{\vv\tran\MA\vv}
\eea
if $\Delta \geq 0$, and zero roots otherwise. Accordingly, when $\Delta \geq 0$, the line intersects the ellipsoid and the entire line segment ${\vy + \alpha \vv: \alpha \in [\alpha_1,\alpha_2]}$ is inside the ellipsoid, hence the shortest distance is zero.

On the other hand, when $\Delta_\alpha < 0$, there is no intersection between the line and the ellipsoid, we seek to find the shortest distance pair by solving the following optimization:
\bea
\min_{\vzz \in \Real{3},\alpha \in \Real{}} & \norm{\vzz - (\vy + \alpha\vv)}^2 \label{eq:minellipsoidline}\\
\subject & (\vzz - \vxx)\tran \MA (\vzz - \vxx) \leq 1
\eea
Similarly, problem~\eqref{eq:minellipsoidline} satisfies LICQ and we write down the KKT conditions:
\bea
(\vzz - \vxx)\tran \MA (\vzz - \vxx) \leq 1 \\
\lambda \geq 0 \\
\nabla_{\vxx} \calL \triangleq 2(\vzz - \vy - \alpha\vv) + 2\lambda\MA(\vzz - \vxx) = \zero \label{eq:calLxvanish1}\\
\nabla_{\alpha} \calL \triangleq 2\vv\tran(\alpha\vv + \vy - \vzz) = 0 \label{eq:calLalphavanish1}\\
\lambda((\vzz-\vxx)\tran\MA(\vzz - \vxx) - 1) = 0.
\eea
Let $\vzz_x \triangleq \vzz - \vxx$, we can simplify the equations above:
\bea
\vzz_x\tran \MA \vzz_x - 1\leq 0 \label{eq:zfeasibleel}\\
\lambda \geq 0 \label{eq:lambdanngel} \\
\parentheses{\lambda\MA + (\eye - \vv\vv\tran)}\vzz_x = (\eye - \vv\vv\tran)\vy_x \label{eq:gradientcalLvanishel}\\
\lambda(\vzz_x\tran\MA\vzz_x - 1) = 0, \label{eq:complementarityel}
\eea
where we have combined~\eqref{eq:calLxvanish1} and~\eqref{eq:calLalphavanish1} by first obtaining:
\bea
\alpha = \vv\tran (\vzz_x - \vy_x),
\eea
from~\eqref{eq:calLalphavanish1} and then inserting it to~\eqref{eq:calLxvanish1}. Now we can discuss two cases for the KKT conditions~\eqref{eq:zfeasibleel}-\eqref{eq:complementarityel}. (i) If $\lambda = 0$, then eq.~\eqref{eq:gradientcalLvanishel} reads:
\bea
(\eye-\vv\vv\tran)(\vzz - \vy) = \zero,
\eea
which indicates that either $\vzz = \vy$ or $\vzz - \vy = k \vv$ for some $k \neq 0$ (note that $\vv$ is the eigenvector of $\eye - \vv\vv\tran$ with associated eigenvalue $0$), which both mean that $\vzz$ lies on the line $L(\vy,\vv)$. This is in contradiction with the assumption that there is no intersection between the line and the ellipsoid. (ii) Therefore, $\lambda > 0$ and $\vzz_x\tran\MA\vzz_x = 1$. In this case, we write $\MV \triangleq \eye - \vv\vv\tran \succeq 0$. Since $\lambda > 0$, $\MA \succ 0$, we have $\lambda\MA +\MV \succ 0$ is invertible, and we get from~\eqref{eq:gradientcalLvanishel} that:
\bea
\vzz_x(\lambda) = (\lambda\MA + \MV)\inv \MV\vy_x.
\eea
Substituting it back to $\vzz_x\tran\MA\vzz_x = 1$, we have that $\lambda$ must satisfy:
\bea
g(\lambda) \triangleq \vzz_x(\lambda)\tran \MA \vzz_x(\lambda) -1 = 0.
\eea
To count the number of roots of $g(\lambda)$ within $\lambda > 0$, we note that:
\bea
g(\lambda \rightarrow 0_{+}) = \vy_x\tran \MA \vy_x - 1> 0 \label{eq:glambdazeroel}\\
g(\lambda \rightarrow +\infty) = -1 < 0 \label{eq:glambdainftyel}
\eea
where $\vy_x\tran \MA \vy_x > 1$ because there is no intersection between the line and the ellipsoid and $\vy$ must lie outside the ellipsoid. We then compute the derivative of $g(\lambda)$~\wrt~$\lambda$:
\bea
g'(\lambda) = 2\vzz_x'(\lambda)\tran \MA \vzz_x(\lambda) \nonumber \\
= -2 \vzz_x(\lambda)\tran \MA (\lambda \MA + \MV)\inv \MA \vzz_x(\lambda) < 0 \label{eq:gdotlambdael}
\eea
where $\vzz_x'(\lambda) = -(\lambda\MA + \MV)\inv \MA \vzz_x(\lambda)$ can be obtained by differentiating both sides of eq.~\eqref{eq:gradientcalLvanishel}~\wrt~$\lambda$. Eqs.~\eqref{eq:glambdazeroel}-\eqref{eq:gdotlambdael} show that $g(\lambda)$ is a monotonically decreasing function in $\lambda > 0$, and a unique root exists in the range $\lambda > 0$. Finally, problem~\eqref{eq:minellipsoidline} admits a global minimizer due to positive definiteness of the Hessian of the Lagrangian. 

The proof above suggests that we can also use Newton's root finding algorithm as in~\eqref{eq:newtonrootfinding} to compute the root of $g(\lambda)$. To make sure $g'(\lambda)$ (eq.~\eqref{eq:gdotlambdael}) is well defined at $\lambda_0$, we initialize $\lambda_0 = 10^{-6}$ instead of $\lambda_0 = 0$ in the $\pe$ case.
\end{proof}
\section{Proof of Lemma~\ref{lemma:potentialenergy}}
\label{sec:app:proof-potentialenergy}
\begin{proof}
Let $(\ux_i,\uy_i) \in \pair{\MT \transform X_i, Y_i}$ be the two endpoints of the shortest distance pair, we have that the cost function of~\eqref{eq:primitivealignment} is $\sum_{i=1}^N \| \ux_i - \uy_i \| ^2$. On the other hand, the potential energy of the system is stored in the virtual springs as $\sum_{i=1}^N \frac{k}{2} \| \ux_i - \uy_i\|^2$, which equates the cost if $k=2$. 
\end{proof}
\section{Proof of Theorem~\ref{thm:globalconvergence}}
\label{app:sec:proof-convergence}
\begin{proof}
Let $\calX = \{ P(\vxx_i) \}_{i=1}^N$ and $\calY = \{ P(\vy_i) \}_{i=1}^N$ be two sets of 3D points, and with slight abuse of notation, we will use $\vxx_i \in \Real{3}$ and $\vy_i \in \Real{3}$ to denote the 3D points and their coordinates interchangeably. Under this setup of Example~\ref{ex:pointcloudregistration}, problem~\eqref{eq:primitivealignment} becomes
\bea
\min_{\MR \in \SOthree, \vt \in \Real{3}} \sum_{i=1}^N \norm{\vy_i - \MR \vxx_i - \vt}^2. \label{eq:pointcloudregistrationform}
\eea
Let
\bea
\barx = \frac{1}{N} \sum_{i=1}^N \vxx_i, \xref{i} = \vxx_i - \barx, \MJ = - m \sum_{i=1}^N \hatmap{\xref{i}}^2 \label{eq:xrefbarxJ}
\eea
be the (initial) center of mass of $\calX$, relative positions of $\vxx_i$ \wrt~center of mass, and moment of inertia of $\calX$. According to eq.~\eqref{eq:totalforce}, the total external force is
\bea
\hspace{-4mm} \vf_i' = k(\vy_i - \MR_q \xref{i} - \xcm) - \mu m (\vcm + \MR_q(\vomega \times \xref{i})) \nonumber \\
\hspace{-4mm} \vf = \sum_{i=1}^N \vf_i', \label{eq:totalforcepcr}
\eea
where $k>0$ is the constant spring coefficient. Similarly, according to eq.~\eqref{eq:totaltorque}, the total torque in the body frame is
\bea
\vtau = \sum_{i=1}^N \xref{i} \times (\MR_q\tran \vf_i'). \label{eq:totaltorquepcr}
\eea 
Now we analyze how many equilibrium points eq.~\eqref{eq:Nprimitivedynamics} has. Towards this goal, setting the first two equations of~\eqref{eq:Nprimitivedynamics} to zero, we get
\bea
\vcm = \zero,\quad \vomega = \zero, \label{eq:zerolinearangularvel}
\eea
which implies that the system must have zero linear velocity and angular velocity at equilibrium. Substituting~\eqref{eq:zerolinearangularvel} to the force and torque expressions in~\eqref{eq:totalforcepcr} and~\eqref{eq:totaltorquepcr}, we have
\bea
\vf_i' = k(\vy_i - \MR_q \xref{i} - \xcm), \\
\vf = k \sum_{i=1}^N \vy_i - \MR_q \xref{i} - \xcm, \label{eq:totalforceequilibrium} \\
\vtau = k \sum_{i=1}^N \xref{i} \times (\MR_q\tran (\vy_i - \MR_q \xref{i} - \xcm)) \nonumber \\
= k \sum_{i=1}^N \xref{i} \times \MR_q\tran (\vy_i - \xcm), \label{eq:totaltorqueequilibrium}
\eea
where we have used the equality that $\xref{i} \times \xref{i} = \zero$. Now we set the last two equations of~\eqref{eq:Nprimitivedynamics} to zero (\ie,~the system has no linear or angular acceleration), we have that 
\bea
\vf = \zero, \quad \vtau = \zero,
\eea
which implies that external forces and torques must balance at an equilibrium point. From $\vf=\zero$ and the expression of $\vf$ in~\eqref{eq:totalforceequilibrium}, we obtain
\bea
\sum_{i=1}^N \vy_i - \MR_q\xref{i} - \xcm = \zero \Longrightarrow \\
\boxed{\xcm = \frac{1}{N} \parentheses{\sum_{i=1}^N \vy_i - \MR_q \xref{i}} = \frac{1}{N} \sum_{i=1}^N\vy_i := \bary}, \label{eq:xcmsolution}
\eea
where we have used
\bea
\frac{1}{N}\sum_{i=1}^N \MR_q \xref{i} = \frac{1}{N} \MR_q \sum_{i=1}^N \xref{i} = \zero
\eea
from the definition of $\xref{i}$ in~\eqref{eq:xrefbarxJ}. Eq.~\eqref{eq:xcmsolution} states that $\calX$ and $\calY$ must have their center of mass aligned at an equilibrium point. Now using a similar notation $\yref{i} \triangleq \vy_i - \bary$, $\vtau=\zero$ from eq.~\eqref{eq:totaltorqueequilibrium} implies that
\bea
\boxed{\sum_{i=1}^N \xref{i} \times \MR_q\tran \yref{i} = \zero}. \label{eq:equivalentequilibrium}
\eea
Eq.~\eqref{eq:xcmsolution} and~\eqref{eq:equivalentequilibrium} are the necessary and sufficient condition for an equilibrium point $\dstate = \zero$. Now we are ready to prove the four claims in Theorem~\ref{thm:globalconvergence}. We first prove~\ref{item:optimalsolution}.

{\bf\ref{item:optimalsolution}: Optimal solution is an equilibrium point}. To show the optimal solution of problem~\eqref{eq:pointcloudregistrationform} is an equilibrium point, we will write down its closed-form solution and show that it satisfies~\eqref{eq:xcmsolution} and~\eqref{eq:equivalentequilibrium}. 
\begin{lemma}[Closed-form Point Cloud Registration]
\label{lemma:pcrclosedform}
The global optimal solution to~\eqref{eq:pointcloudregistrationform} is
\bea
\vt^\star = \bary - \MR^\star \barx, \\
\MR^\star = \MUplus\MVplus\tran,
\eea
where $\MUplus,\MVplus \in \SOthree$ are obtained from the singular value decomposition:
\bea
\MM = \sum_{i=1}^N \yref{i}\xref{i}\tran = \MU \MS \MV\tran, \quad \MU, \MV \in \Othree, \\
\MUplus = \MU \diag{[1,1,\det{\MU}]} \in \SOthree, \\
\MVplus = \MV \diag{[1,1,\det{\MV}]} \in \SOthree.
\eea
Using $\MUplus, \MVplus$, we have $\MM = \MUplus \MS' \MVplus\tran$, with
\bea
\MS' = \diag{[s_1,s_2,s_3 \det{\MU\MV}]}.
\eea
\end{lemma}
Lemma~\ref{lemma:pcrclosedform} is a standard result in point cloud registration~\cite{Horn87josa,markley1988jas-svdAttitudeDeter}. Using $(\vtstar,\MRstar)$, one immediately sees that the center of mass of $\calX$ is transformed to
\bea
\xcm = \MRstar \barx + \vtstar = \MRstar \barx + \bary - \MRstar \barx = \bary
\eea
and coincide with $\bary$, hence satisfies eq.~\eqref{eq:xcmsolution}. Now replace $\MR_q$ with $\MRstar = \MUplus \MVplus\tran$ in eq.~\eqref{eq:equivalentequilibrium}, our goal is to show
\bea
\vtau(\MRstar) \triangleq \sum_{i=1}^N \xref{i} \times (\MVplus \MUplus\tran \yref{i})
\eea
equal to zero. Towards this goal, we will show each entry of $\vtau(\MRstar)$ is zero,~\ie,~$\ve_j\tran \vtau(\MRstar) = 0$ for $j=1,2,3$. Note that
\bea
\ve_1\tran \vtau(\MRstar) = \sum_{i=1}^N \ve_1\tran \hatmap{\xref{i}} \MVplus\MUplus\tran \yref{i} \label{eq:ve1vtauRstart}\\
= \sum_{i=1}^N \xref{i}\tran (\ve_2\ve_3\tran - \ve_3\ve_2\tran) \MVplus\MUplus\tran \yref{i} \\
= \sum_{i=1}^N \trace{\MUplus\tran \yref{i} \xref{i}\tran (\ve_2\ve_3\tran - \ve_3\ve_2\tran)\MVplus} \\
= \trace{\MUplus\tran \parentheses{\sum_{i=1}^N \yref{i}\xref{i}\tran } (\ve_2\ve_3\tran - \ve_3\ve_2\tran) \MVplus} \\
= \trace{\MUplus\tran \MM (\ve_2\ve_3\tran - \ve_3\ve_2\tran) \MVplus } \\
= \trace{\MUplus\tran \MUplus \MS' \MVplus\tran (\ve_2\ve_3\tran - \ve_3\ve_2\tran) \MVplus } \\
= \trace{\MVplus \MS' \MVplus\tran (\ve_2\ve_3\tran - \ve_3\ve_2\tran)} \\ 
= [\MVplus \MS' \MVplus\tran]_{32} - [\MVplus \MS' \MVplus\tran]_{23} = 0 \label{eq:ve1vtauRend}
\eea
where the last ``$=0$'' holds because $\MVplus \MS' \MVplus\tran$ is an symmetric matrix, and we have used the fact that
\bea
\ve_1\tran \hatmap{\vxx} \equiv \vxx\tran (\ve_2\ve_3\tran - \ve_3\ve_2\tran), \forall \vxx \in \Real{3}.
\eea
By the same token, one can verify that
\bea
\ve_2\tran \vtau(\MRstar) = [\MVplus \MS' \MVplus\tran]_{13} - [\MVplus \MS' \MVplus\tran]_{31} = 0,\\
\ve_3\tran \vtau(\MRstar) = [\MVplus \MS' \MVplus\tran]_{21} - [\MVplus \MS' \MVplus\tran]_{12} = 0.
\eea
Therefore, the optimal solution $(\vtstar,\MRstar)$ is an equilibrium point of the system~\eqref{eq:Nprimitivedynamics}.

{\bf\ref{item:foursolution} and~\ref{item:threebypi}: Three spurious equilibrium points}. We now show that besides the optimal equilibrium point $(\vtstar,\MRstar)$, the equation~\eqref{eq:equivalentequilibrium} has three and only three different solutions if $s_1 > s_2 > s_3 > 0$, which we denote as \emph{generic configuration}. Towards this goal, let us assume there is a rotation matrix $\MR_q$ that satisfies~\eqref{eq:equivalentequilibrium}, and we write it as
\bea
\MR_q = \MUplus \barMR \MVplus\tran.
\eea 
Note that such a parametrization is always possible with
\bea
\barMR = \MUplus\tran \MR_q \MVplus \in \SOthree.
\eea
Using this parametrization, $\vtau(\MR_q) = \zero$ is equivalent to
\bea
\MZ \triangleq \MVplus\barMR \MS' \MVplus\tran
\eea
being symmetric (using similar derivations as in~\eqref{eq:ve1vtauRstart}-\eqref{eq:ve1vtauRend}). Then it is easy to see that $\MZ$ being symmetric is equivalent to $\barMR \MS'$ being symmetric because $\barMR\MS' = \MVplus\tran \MZ \MVplus$. Explicitly, we require
\bea
\barMR \MS' = (\barMR\MS')\tran = \MS' \barMR\tran. 
\eea
Since $s_1>s_2>s_3 >0$, $\MS'$ is invertible and $(\MS')\inv = \diag{[1/s_1,1/s_2,1/s_3']}$ with $s_3' = s_3 \det{\MU\MV}$. Therefore, $\barMR \MS' = \MS' \barMR\tran$ is equivalent to
\bea
\hspace{-4mm} \barMR = \MS' \barMR\tran (\MS')\inv \Leftrightarrow \\
\underbrace{\bmat{ccc}
r_{11} & r_{12} & r_{13} \\
r_{21} & r_{22} & r_{23} \\
r_{31} & r_{32} & r_{33}
\emat}_{\barMR \in \SOthree} = 
\underbrace{\bmat{ccc}
r_{11} & \frac{s_1}{s_2} r_{21} & \frac{s_1}{s_3'} r_{31} \\
\frac{s_2}{s_1} r_{12} & r_{22} & \frac{s_2}{s_3'} r_{32} \\
\frac{s_3'}{s_1} r_{13} &  \frac{s_3'}{s_2}r_{23} & r_{33}
\emat}_{\MS' \barMR\tran (\MS')\inv}. \label{eq:barMRSO3}
\eea
Now we use $\abs{\frac{s_1}{s_2}}, \abs{\frac{s_1}{s_3'}}, \abs{\frac{s_2}{s_3'}}>1$, and the fact that both sides of~\eqref{eq:barMRSO3} are rotation matrices:
\bea
\hspace{-4mm} r_{11}^2 + r_{21}^2 + r_{31}^2 = r_{11}^2 + \parentheses{\frac{s_1}{s_2}}^2 r_{21}^2 + \parentheses{\frac{s_1}{s_3'}}^2 r_{31}^2 = 1, \\
\hspace{-4mm} r_{33}^2 + r_{32}^2 + r_{31}^2 = r_{33}^2 + \parentheses{\frac{s_2}{s_3'}}^2 r_{32}^2 + \parentheses{\frac{s_1}{s_3'}}^2 r_{31}^2 = 1,
\eea
which implies that
\bea
\parentheses{\parentheses{\frac{s_1}{s_2}}^2-1} r_{21}^2 + \parentheses{ \parentheses{\frac{s_1}{s_3'}}^2 - 1}r_{31}^2 = 0,\\
\parentheses{\parentheses{\frac{s_2}{s_3'}}^2-1} r_{32}^2 + \parentheses{\parentheses{\frac{s_1}{s_3'}}^2 - 1} r_{31}^2 = 0,
\eea
and hence $r_{21} = r_{31} = r_{32} = 0$. Substituting them back into~\eqref{eq:barMRSO3}, we have $r_{12}=r_{13}=r_{23} = 0$. Therefore, we conclude that $\barMR$ is a diagonal matrix. However, there are only four rotation matrices that are diagonal:
\bea
\barMR_1 = \diag{[1,1,1]},\\
\barMR_2 = \diag{[1,-1,-1]},\\
\barMR_3 = \diag{[-1,1,-1]},\\
\barMR_4 = \diag{[-1,-1,1]}.
\eea
As a result, the equation~\eqref{eq:equivalentequilibrium} has four and only four solutions. Note that $\barMR_1 = \eye_3$ corresponds to the optimal equilibrium point $\MRstar$, and the angular distance between $\MRstar$ and the other three spurious equilibrium points $\MUplus \barMR_j \MVplus\tran, j=2,3,4$ is:
\bea
\abs{\arccos\parentheses{\frac{\trace{(\MRstar)\tran \MUplus \barMR_j \MVplus\tran}-1}{2}}} \nonumber \\
= \abs{\arccos\parentheses{  \frac{ \trace{\MVplus\MUplus\tran \MUplus \barMR_j \MVplus\tran}  -1}{2}  } } \nonumber \\
= \abs{\arccos\parentheses{  \frac{\trace{\barMR_j} - 1}{2} } } = \pi.
\eea

{\bf\ref{item:localunstable}: Locally unstable spurious equilibrium points}. Lastly, we are ready to show that the three spurious equilibrium points are locally unstable. Let the system be at one of the three spurious equilibrium points $\state = (\bary,\MUplus \barMR_j \MVplus\tran,\zero,\zero),j=2,3,4$ with zero translational and angular velocities (such that the total energy of the system equals the total potential energy of the system due to zero kinetic energy), and consider a small perturbation to the equilibrium point:
\bea
\state_{\Delta} = (\bary,\MR_{\Delta} \MUplus \barMR_j \MVplus\tran,\zero,\zero),
\eea
with a perturbing rotation $\deltaR \in \SOthree$. The total (potential) energy of the system at $\state$ is
\bea
V(\state) = \frac{k}{2} \sum_{i=1}^N \norm{\vy_i - \MUplus \barMR_j \MVplus\tran \xref{i} - \bary}^2 \\
= \frac{k}{2} \sum_{i=1}^N \norm{\yref{i} -  \MUplus \barMR_j \MVplus\tran \xref{i}}^2 \\
= \overbrace{\frac{k}{2} \sum_{i=1}^N \norm{\yref{i}}^2 + \frac{k}{2}\sum_{i=1}^N \norm{\xref{i}}^2}^{:=E} - \nonumber \\
k \sum_{i=1}^N \trace{\xref{i}\tran \MVplus\barMR_j\MUplus\tran\yref{i}} \\
= E - k \trace{\parentheses{\sum_{i=1}^N\yref{i}\xref{i}\tran} \MVplus \barMR_j \MUplus\tran} \\
= E - k\trace{\MUplus \MS' \MVplus\tran \MVplus \barMR_j \MUplus\tran} \\
= E - k\trace{ \MS' \barMR_j}.
\eea
The total energy of the system at $\deltastate$ is
\bea
V(\deltastate) = \frac{k}{2} \sum_{i=1}^N \norm{\yref{i} - \deltaR \MUplus \barMR_j \MVplus\tran \xref{i}}^2 \\
= E - k \sum_{i=1}^N \trace{\xref{i}\tran \MVplus \barMR_j \MUplus\tran \deltaR\tran\yref{i}} \\
= E - k \trace{\parentheses{\sum_{i=1}^N \yref{i} \xref{i}\tran} \MVplus\barMR_j \MUplus\tran \deltaR\tran} \\
= E - k \trace{\MUplus \MS' \barMR_j \MUplus\tran \deltaR\tran} \\
= E - k \trace{\MS'\barMR_j \MUplus\tran \deltaR\tran \MUplus}.
\eea
Therefore, we have that the difference of energy from $V(\state)$ to $V(\deltastate)$ is
\bea
V(\state) - V(\deltastate) = k \inner{\MS' \barMR_j}{\MUplus\tran \deltaR \MUplus - \eye_3}.
\eea
Using the Rodrigues’ rotation formula on $\deltaR$:
\bea
\deltaR = \cos\theta \eye_3 + \sin\theta \hatmap{\vu} + (1-\cos\theta)\vu\vu\tran,
\eea
where $\theta$ is the rotation angle and $\vu \in \usphere{2}$ is the rotation axis, we have
\bea
\MUplus\tran \deltaR \MUplus - \eye_3 = \nonumber \\
(\cos\theta -1)\eye_3 + (1-\cos\theta)\MUplus\tran\vu\vu\tran\MUplus + \nonumber \\ 
\sin\theta \MUplus\tran\hatmap{\vu}\MUplus,
\eea
and the last term $\sin\theta \MUplus\tran\hatmap{\vu}\MUplus$ is skew-symmetric. Since $\MS'\barMR_j$ is diagonal (and its inner product with any skew-symmetric matrix is zero), we have
\bea
\hspace{-4mm} V(\state) - V(\deltastate) = k(1-\cos\theta) \inner{\MS'\barMR_j}{\vz\vz\tran - \eye_3} \\
= k(1-\cos\theta) \parentheses{ \vz\tran (\MS'\barMR_j)\vz - \trace{\MS'\barMR_j} },
\eea
where we have denoted $\vz \triangleq \MUplus\tran\vu \in \usphere{2}$. Now using the expression for $\barMR_j, j=2,3,4$, we have:
\bea
\MS'\barMR_2 = \diag{[s_1,-s_2,-s_3']},\\
\MS'\barMR_3 = \diag{[-s_1,s_2,-s_3']}, \\
\MS'\barMR_4 = \diag{[-s_1,-s_2,s_3']}.
\eea 
Hence, when $j=2$, we choose $\vz = [1,0,0]\tran$, so that
\bea
V(\state) - V(\deltastate) = k(1-\cos\theta)(s_2 + s_3') > 0;
\eea
when $j=3$, we choose $\vz=[0,1,0]\tran$, so that
\bea
V(\state) - V(\deltastate) = k(1-\cos\theta)(s_1 + s_3') > 0;
\eea
when $j=4$, we choose $\vz = [0,0,1]\tran$, so that
\bea
V(\state) - V(\deltastate) = k(1-\cos\theta)(s_1 + s_2) > 0.
\eea
This implies that, in all three cases, there exist small rotational perturbations with angle $\theta$ along axis $\MUplus\vz$ (recall that $\vz = \MUplus\tran \vu$), such that this small perturbation will cause a strict decrease in the total energy of the system. As a result, the system is locally unstable at the three spurious equilibrium points. Using Lyapunov's local stability theory~\cite{Slotine91book-nonlinearcontrol}, we know that, unless starting exactly at one of the spurious equilibrium points, the system will never converge to these locally unstable equilibrium points. 
\end{proof}
\section{Corner Cases of Point Cloud Registration}
\label{sec:app:cornercases}

We show two examples of corner cases of point cloud registration where the configuration is not general and violates the $s_1>s_2>s_3>0$ assumption in Section~\ref{app:sec:proof-convergence}, they correspond to when there is no noise between $\calX$ and $\calY$ and both of them have symmetry.

When $N=3$ (Fig.~\ref{fig:symmetry}(a)), consider both $\calX$ (blue) and $\calY$ (red) are equilateral triangles with $l$ being the length from the vertex to the center. Assume the particles have equal masses such that the CM is also the geometric center $O$, and all virtual springs have equal coefficients. $\calX$ is obtained from $\calY$ by first rotating counter-clockwise (CCW) around $O$ with angle $\theta$, and then flipped about the line that goes through point 1 and the middle point between point 2 and 3. We will show that this is an equilibrium point of the dynamical system for any $\theta$. When the CM of $\calX$ and the CM of $\calY$ aligns, we know the forces $\vf_i,i=1,2,3$ are already balanced. It remains to show that the torques $\vtau_i,i=1,2,3$ are also balanced for any $\theta$. $\vtau_1$ and $\vtau_3$ applies clockwise (CW, cyan) and the value of their sum is:
\bea
\hspace{-10mm} \| \vtau_1 + \vtau_3 \| = \| \vtau_1 \| + \| \vtau_3 \| \\
= kl^2 \left( \sin \theta +  \sin\beta \right) \\
= kl^2 \left( \sin\theta + \sin\left(\theta + \frac{2\pi}{3} \right) \right) \\
= kl^2 \sin\left( \theta + \frac{\pi}{3} \right),
\eea
and $\tau_2$ applies CCW (green) and its value is:
\bea
\| \vtau_2 \| = kl^2 \sin\alpha  = kl^2 \sin\left( \frac{2\pi}{3} - \theta \right) \\
= kl^2 \sin\left( \theta + \frac{\pi}{3} \right).
\eea
Therefore, the torques cancel with each other and the configuration in Fig.~\ref{fig:symmetry}(a) is an equilibrium state for all $\theta$. However, it is easy to observe that this type of equilibrium is unstable because any perturbation that drives point 2 out of the 2D plane will immediately drives the system out of this type of equilibrium. When $N=4$, one can verify that same torque cancellation happens: 
\bea
\| \vtau_1 \| = kl^2 \sin\beta = kl^2 \sin\left( \theta + \frac{\pi}{2} \right) \\
=  kl^2 \sin\left( \frac{\pi}{2} - \theta \right) = kl^2 \sin\alpha = \| \vtau_3 \|,
\eea
and the system also has infinite locally unstable equilibria.


\newcommand{\mpwthree}{6cm}
\newcommand{\myhspace}{\hspace{-3mm}}

\begin{figure}[t]
	\begin{center}
	\begin{minipage}{\textwidth}
	\begin{tabular}{c}%
			\begin{minipage}{\mpwthree}%
			\centering%
			\includegraphics[width=0.9\columnwidth]{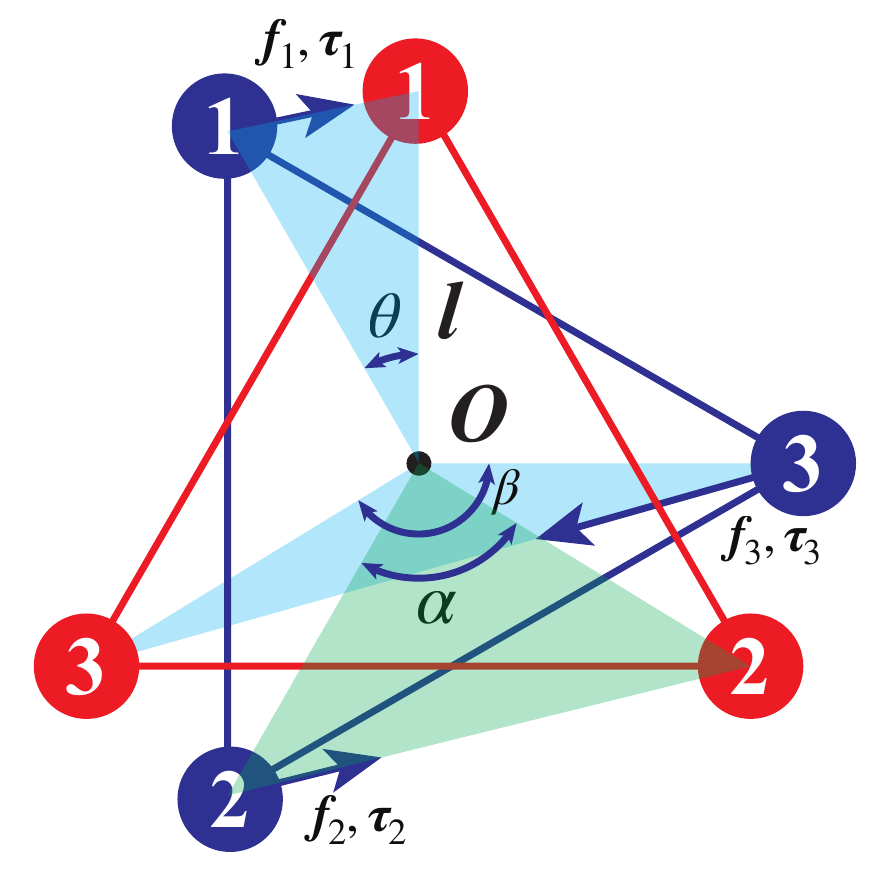} \\
			{\small (a) Equilateral triangle.}
			\end{minipage}
		\\ \hspace{6mm}
			\begin{minipage}{\mpwthree}%
			\centering%
			\includegraphics[width=0.9\columnwidth]{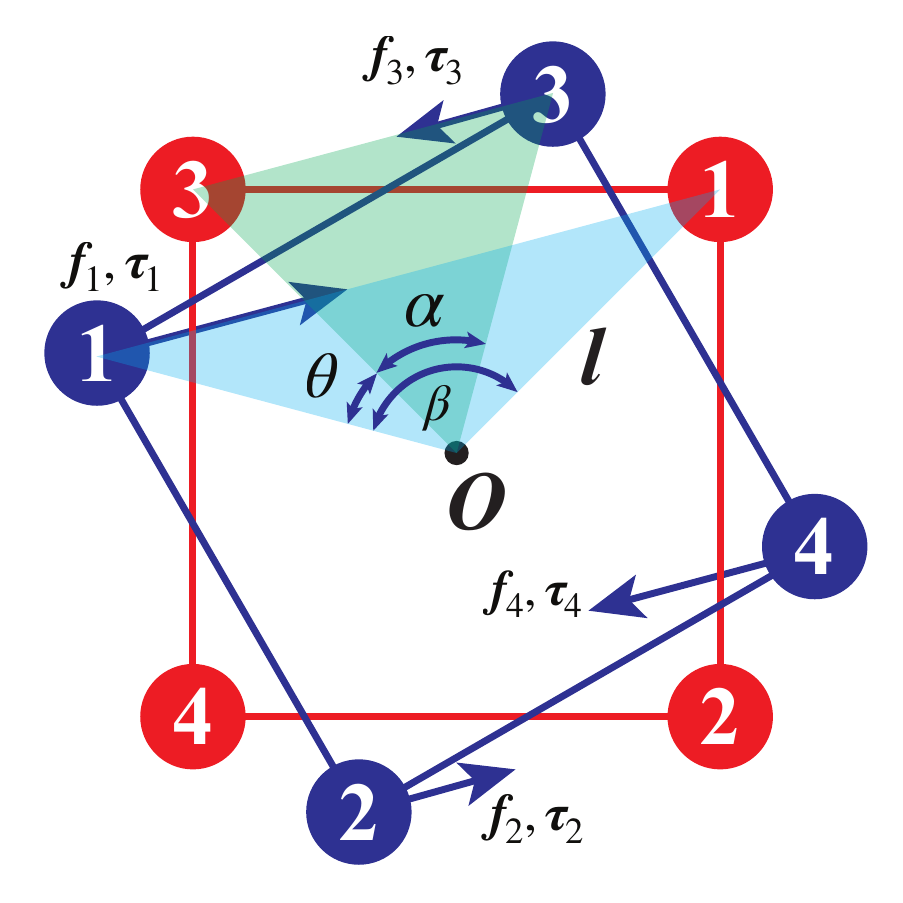} \\
			{\small (b) Square.}
			\end{minipage}
	\end{tabular}
	\end{minipage} 
	\caption{Examples of symmetric point clouds: (a) an equilateral triangle and (b) a square. The dynamical system has infinite equilibrium points.
	\label{fig:symmetry}} 
	\vspace{-5mm} 
	\end{center}
\end{figure}
\section{Extra Experimental Results}
\label{sec:app:experiments}
\renewcommand{\mpwsingletwo}{4.5cm}
\begin{figure}[h]
\begin{minipage}{\textwidth}
\begin{tabular}{cc}%
\hspace{-6mm}\begin{minipage}{\mpwsingletwo}%
\centering%
\includegraphics[width=\columnwidth]{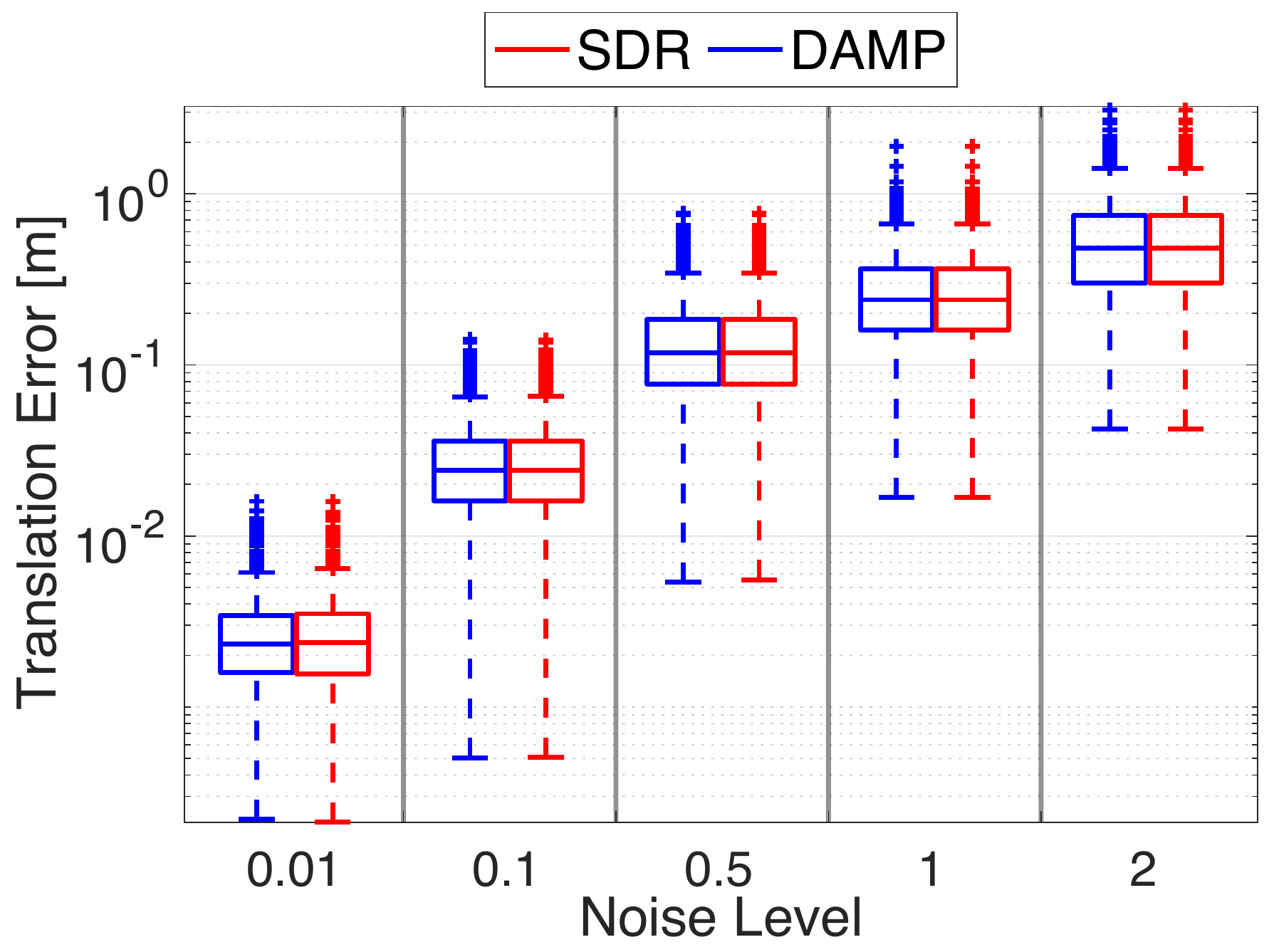}
\end{minipage}
&
\hspace{-5mm} \begin{minipage}{\mpwsingletwo}%
\centering%
\includegraphics[width=\columnwidth]{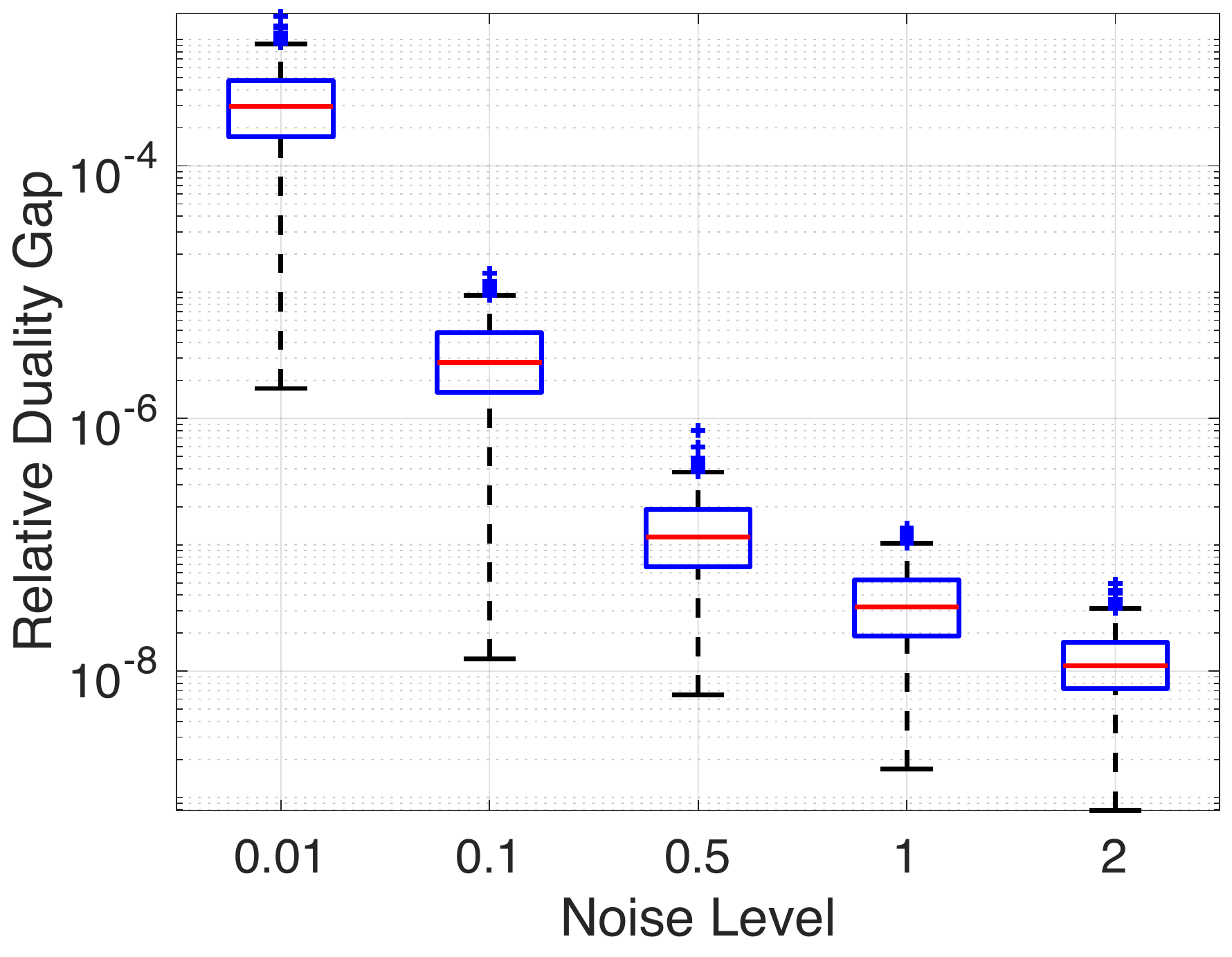}
\end{minipage}
\end{tabular}
\end{minipage}
\vspace{-3mm}
\caption{Translation estimation error of \nameshort compared with the certifiably optimal \sdr solver~\cite{Briales17cvpr-registration} on random primitive registration with increasing noise levels. Right plot shows the relative duality gap computed from \sdr, which certifies that both \nameshort and \sdr attains the globally optimal solution.}
\label{fig:app-mesh-t-duality}
\end{figure}

\newcommand{\mpwdoublethree}{5.6cm}
\begin{figure*}[h]
\begin{minipage}{\textwidth}
\centering
\begin{tabular}{ccc}%
\hspace{-4mm}\begin{minipage}{\mpwdoublethree}%
\centering%
\includegraphics[width=\columnwidth]{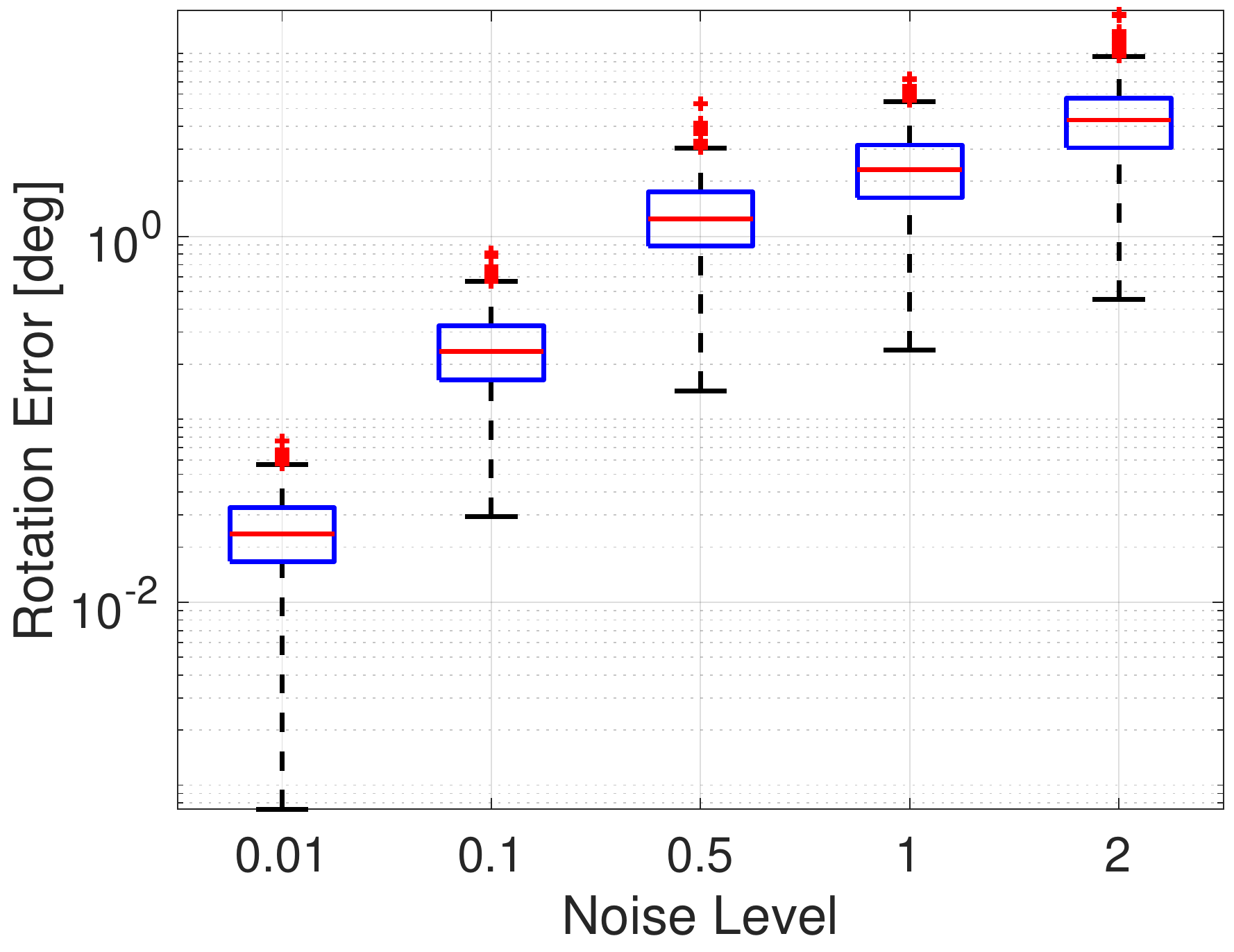}
\end{minipage}
&
\hspace{-2mm} \begin{minipage}{\mpwdoublethree}%
\centering%
\includegraphics[width=\columnwidth]{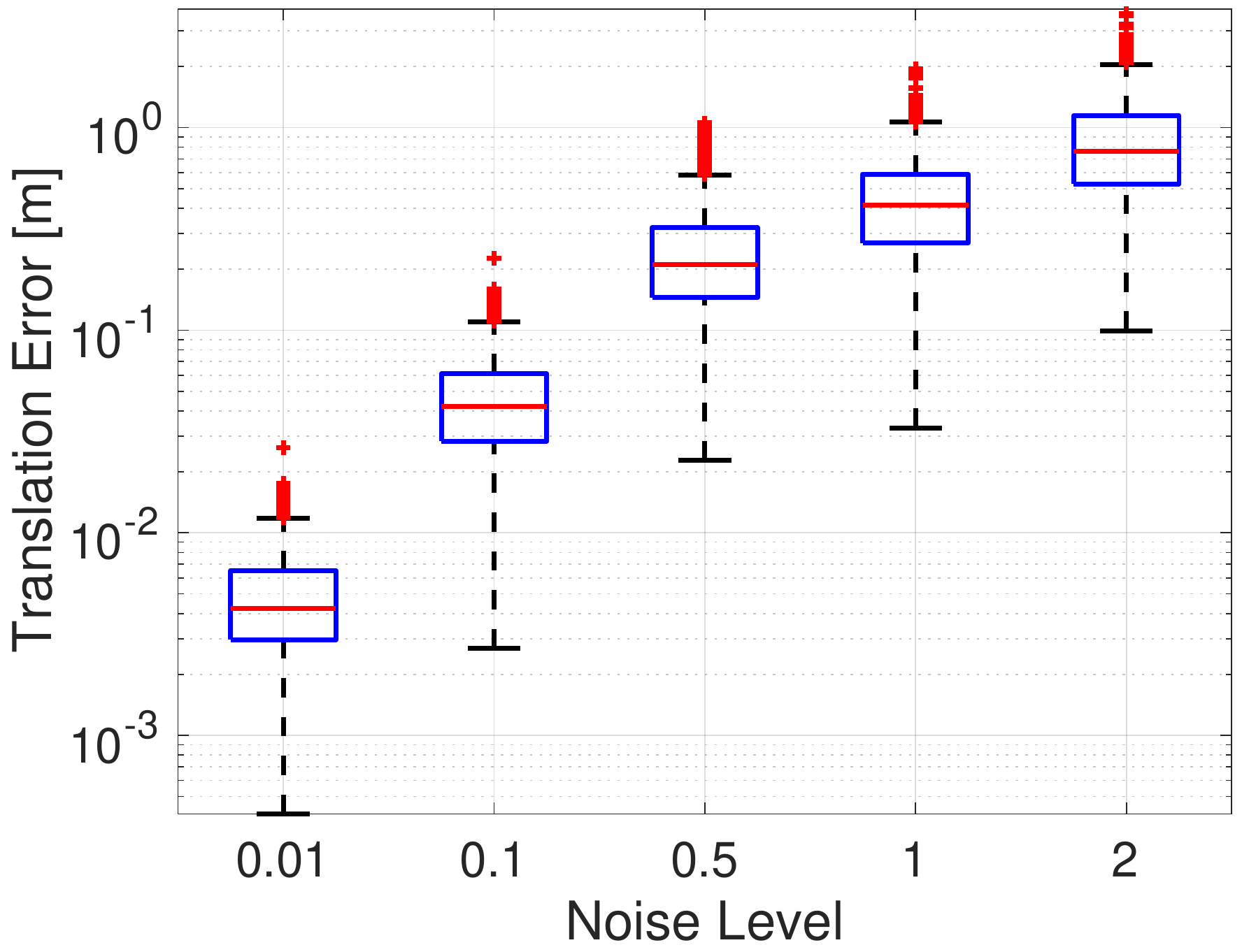}
\end{minipage}
&
\hspace{-2mm} \begin{minipage}{\mpwdoublethree}%
\centering%
\includegraphics[width=\columnwidth]{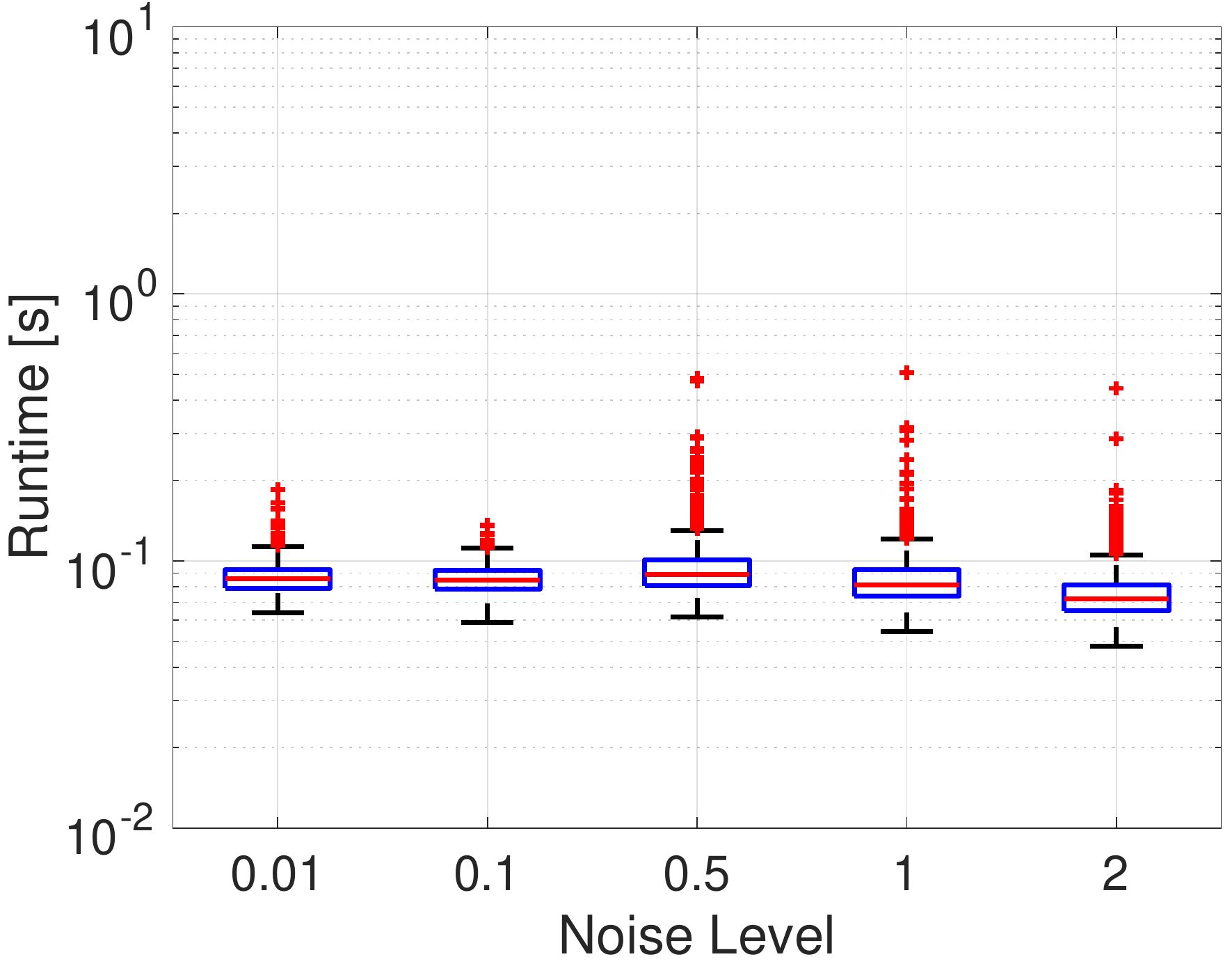}
\end{minipage}

\end{tabular}
\end{minipage}
\vspace{-3mm}
\caption{Rotation error, translation error and runtime of \nameshort on aligning point cloud observation to the robot primitive model (\cf~Fig.~\ref{fig:all-apps}(b) in main text) under increasing noise levels. Although there is no guaranteed globally optimal solver to verify \nameshort's optimality, the accurate estimations strongly indicate \nameshort's global convergence (1000 Monte Carlo runs per noise level).}
\label{fig:app-robot-primitive}
\end{figure*}

{\bf Mesh registration}. Fig.~\ref{fig:app-mesh-t-duality} shows the translation error of \nameshort compared with \sdr~\cite{Briales17cvpr-registration} on varying noise levels, as well as the relative duality gap of \sdr. Because the relative duality gap of \sdr is numerically zero, we can say that \sdr finds the globally optimal solutions in all Monte Carlo runs. Then we look at the translation error boxplot and observe that \nameshort always returns the same solution as \sdr, which indicates that \nameshort always converges to the optimal solution. 

{\bf Robot primitive registration}. Fig.~\ref{fig:app-robot-primitive} plots the rotation error, translation error and runtime of \nameshort on registering a noisy point cloud observation to the robot primitive including planes, spheres, cylinders and cones, under increasing noise levels, where 1000 Monte Carlo runs are performed at each noise level. We find that \nameshort always returns an accurate pose estimation, even when the noise standard deviation is 2 (note that the scene radius is 10), strongly suggesting that \nameshort always converges to the optimal solution. Moreover, \nameshort has a runtime that is below 1 second (recall that our implementation is in Matlab with for loops, because \nameshort is a general algorithm that checks the type of the primitive for each correspondence).

{\small
\bibliographystyle{ieee_fullname}
\bibliography{references}
}

\end{document}